\documentclass[11pt]{article}

\usepackage{acl}
\usepackage{times}
\usepackage{latexsym}
\usepackage[utf8]{inputenc}
\usepackage[T1]{fontenc}
\usepackage{microtype}
\usepackage{inconsolata}
\usepackage{graphicx}
\usepackage{wrapfig}
\usepackage{subcaption}
\usepackage{booktabs}

\hypersetup{
  colorlinks=true,
  linkcolor=blue,
  citecolor=blue,
  urlcolor=blue,
  pdftitle={STAGE: A Full-Screenplay Benchmark for Reasoning over Evolving Stories},
  pdfauthor={Qiuyu Tian, Yiding Li, Yingce Xia, Fengyi Chen, Youyong Kong, Fan Guo, Yuyao Li, Jinjing Shen, Zhijing Xie, Yiyun Luo, Xin Zhang, Zequn Liu}
}

\usepackage{placeins}
\usepackage{amsmath,amssymb}
\usepackage{multirow}
\usepackage{makecell}
\usepackage{xcolor}
\usepackage{tcolorbox}
\usepackage{enumitem}
\usepackage{algorithm}
\usepackage{algorithmic}
\usepackage{dblfloatfix}

\usepackage[english]{babel}

\newcommand{\stageTableFont}{\small}

\newtcolorbox{stageprompt}[1]{
  colback=black!2,
  colframe=black!45,
  colbacktitle=black!7,
  coltitle=black,
  boxrule=0.45pt,
  arc=1pt,
  left=5pt,
  right=5pt,
  top=4pt,
  bottom=4pt,
  toptitle=3pt,
  bottomtitle=3pt,
  fonttitle=\bfseries\small,
  fontupper=\small,
  before skip=6pt,
  after skip=6pt,
  title={#1}
}
\newcommand{\stagefield}[1]{\texttt{\{#1\}}}

\title{STAGE: A Full-Screenplay Benchmark for Reasoning over Evolving Stories}

\author{%
\begin{tabular}{@{}*{4}{>{\centering\arraybackslash}p{0.22\textwidth}}@{}}
\small
Qiuyu Tian$^{1,2}$ & Yiding Li$^{5}$ & Yingce Xia$^{2}$ & Fengyi Chen$^{3}$ \\
Youyong Kong$^{1}$ & Fan Guo$^{5}$ & Yuyao Li$^{5}$ & Jinjing Shen$^{5}$ \\
Zhijing Xie$^{5}$ & Yiyun Luo$^{5}$ & Xin Zhang$^{5}$ & Zequn Liu$^{2,*}$ \\[0.45em]
\multicolumn{4}{c}{%
\begin{tabular}{c}
\small $^{1}$Southeast University, Nanjing, China \\
\small $^{2}$Beijing Zhongguancun Academy, Beijing, China \\
\small $^{3}$Nanjing Normal University, Nanjing, China \\
\small $^{5}$ZhuiWen Technology Co., Ltd., Beijing, China \\
\small $^{*}$Corresponding author.
\end{tabular}}
\end{tabular}}

\begin{document}

\maketitle

\begin{abstract}
Movie screenplays are a demanding testbed for long-form narrative
understanding, as characters' goals, beliefs, knowledge, and relationships
evolve continuously across scenes. However, existing benchmarks primarily
evaluate isolated facts from the completed screenplay, leaving unassessed
whether models can track the evolving state of characters as the story unfolds.
We introduce STAGE, a benchmark over 151 English and Chinese full-length
screenplays, built on a provenance-linked narrative backbone that recovers the
state and epistemic access of each character at every point along its timeline.
Three tasks derived from the backbone jointly probe whether models can maintain,
explain, and act on evolving narrative state: \textit{Character Development
Tracking} updates a focal character's state between checkpoints,
\textit{Cross-Scene Narrative Evolution Reasoning} targets cross-scene state
transitions, and \textit{In-Script Character Role-Playing} requires responses
bounded by the character's state and knowledge at a specified point. We identify
three failure modes of current LLMs: silent forgetting under recursive state
updating, limited cross-scene reasoning even when all relevant evidence is
supplied, and a trade-off in role-playing where stylistic character fidelity and
screenplay-grounded memory faithfulness are optimized by different memory-access
strategies. STAGE thus provides a unified framework for diagnosing how current
models fail to track, reason about, and enact story evolution. Benchmark assets
and evaluation code are available at
\url{https://github.com/roytian1992/STAGE_v0}.
\end{abstract}

\section{Introduction}

A screenplay presents an ordered sequence of state changes. As the script
proceeds, a character's goals are revised, beliefs are
corrected, relationships are reconfigured, and constraints are imposed and
lifted. Understanding such a text requires a time-indexed representation of the
story, covering the currently valid state, the cause of each change, the
characters with access to it, and the elements that persist unchanged
\citep{graesser1990quest,graesser1994constructing,riedl2010narrativeplanning}.
A question about an intermediate stage of the story therefore calls for
checkpoint-specific state reconstruction. Retrieval over the completed script
remains insufficient for this purpose, even for long-context
LLMs~\citep{liu2024lost}.
Such state reconstruction is broadly useful in settings with an accumulating
interaction history (e.g., long-horizon agents
\citep{park2023generative,xu2025amem,tan2025reflective}), and the explicit scene
order and textually realized state changes of screenplays make them a natural
starting point for studying it.

Evolving narratives introduce two asymmetries that static document
comprehension often leaves implicit. First, evidence accumulates, but state
validity does not: an earlier goal, belief, relationship, or constraint may be
revised, resolved, or superseded. Second, narrative truth and character
knowledge diverge. An event can alter the story world without being observed by
the focal character, and a later revelation cannot license an earlier action.
Evaluation that ignores these temporal and epistemic boundaries may reward a
factually plausible answer even when it is invalid at the queried point.

\begin{figure*}[t]
    \centering
    \includegraphics[width=\textwidth]{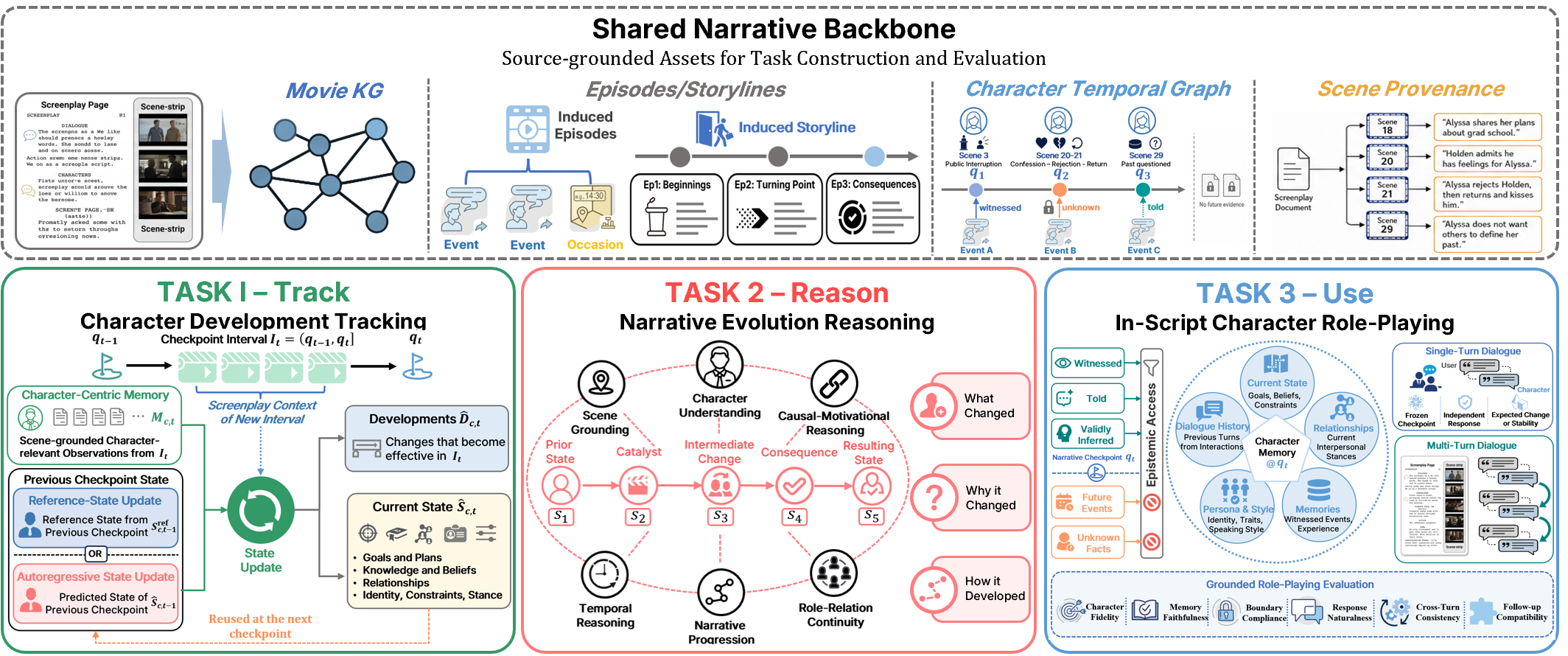}
    \caption{
    Overview of STAGE.
    Ordered screenplay scenes support a source-grounded representation with an
    objective narrative backbone and character-relative temporal views. These
    construction assets remain hidden from evaluated models and define three
    aligned tasks: \textbf{Character Development Tracking} measures successive state
    updates and rollout stability, \textbf{Cross-Scene Narrative Evolution
    Reasoning} evaluates explanations of state change, and \textbf{In-Script
    Character Role-Playing} tests checkpoint-valid generation.
    }
    \label{fig:stage_overview}
\end{figure*}

Existing screenplay and long-form narrative benchmarks primarily emphasize
recognition, summarization, conditioned generation, story QA, or general
long-context reading
\citep{zheng2025stpar,gorinski2015movie,papalampidi2020screenplay,saxena2024moviesum,chen2024hollmwood,mirowski2023dramatron,zheng2025cmlbench}.
Related long-form benchmarks broaden document-level coverage
\citep{kovcisky2018narrativeqa,bai2023longbench,ran2024novelqa,hamilton2025tldm}.
However, they generally treat the completed story as a static source or assess
tasks independently, leaving open whether models can track accumulating state,
explain changes at the correct time, and act within the knowledge available at
that point.
This is also a cross-use alignment problem: a model may recover the right scene
for a question yet fail to carry its consequence into a later state update, or
imitate a character fluently while exceeding that character's acquired
knowledge. Aligning tracking, reasoning, and generation to the same ordered
evidence enables direct evaluation of all three capabilities.

To address this gap, we introduce STAGE (Screenplay Text, Agents, Graphs \&
Evaluation), a full-screenplay benchmark covering 151 English and Chinese
movies. Its three tasks share a temporal axis: \textbf{Character Development
Tracking} updates a focal character between checkpoints and measures recursive
error accumulation; \textbf{Cross-Scene Narrative Evolution Reasoning} tests
evidence-grounded reasoning about change, timing, knowledge, delayed effects,
and persistence; and \textbf{In-Script Character Role-Playing (ICRP)} requires
natural responses bounded by the character's checkpoint-valid state and
knowledge. Together, they test whether models can \emph{track}, \emph{reason
about}, and \emph{use} story evolution. To define these tasks, we need each
character's state and knowledge at every checkpoint, which remain latent in
screenplay text. We therefore construct a narrative backbone for each film,
comprising a source-grounded objective narrative hierarchy and a
character-relative temporal graph, and derive the tasks from it. By recording
evidence, checkpoints, state validity, and epistemic access once per film, the
backbone makes annotation tractable at full-screenplay scale and keeps the
three tasks grounded in the same source, while remaining hidden from evaluated
models.

Our evaluation identifies three complementary failure modes: recursive updates
progressively omit prior developments; cross-scene answers remain unstable even
when retrieval covers all gold evidence; and role-playing can remain fluent
while failing to use checkpoint-valid memory faithfully. Collectively, our
contributions are:
\begin{itemize}[leftmargin=*,noitemsep]
    \item \textbf{An aligned evolving-story benchmark} spanning tracking,
    cross-scene reasoning, and checkpoint-bounded generation over 151 films.
    \item \textbf{Source-grounded construction} that separates objective
    narrative structure from character-relative state and knowledge.
    \item \textbf{A systematic capability analysis} showing recursive state
    error, unstable cross-scene evidence use, and fluent but weakly grounded
    role-playing.
\end{itemize}

\section{Related Work}
\label{sec:related_work_scope}

\textbf{Narrative structure and evolving state.}
Computational narrative theory models comprehension through goals, causality,
plans, and character intentionality across events
\citep{graesser1990quest,graesser1994constructing,riedl2010narrativeplanning}.
It also distinguishes event ordering from higher-level narrative structure:
\citet{piper-etal-2021-narrative} organize the latter as an
event--scene--narrative-level--plotline--plot hierarchy. Event-centric graphs
make parts of this structure explicit
\citep{yan2023narrativegraph}, while consistency studies show that language
models can vary across contexts even when the underlying facts are unchanged
\citep{novikova2025consistency}. STAGE operationalizes this coarse-to-fine
principle for screenplays and adds a character-relative temporal layer. This
supports checkpointed evaluation that separates objective events from
character access, and current states from the developments that produced them.

\textbf{Long-form benchmarks and retrieval.}
NarrativeQA, BookQA, LongBench, NovelQA, and TLDM evaluate retrieval and
document-level comprehension across books, scripts, and mixed long documents
\citep{kovcisky2018narrativeqa,kozik2020bookqa,bai2023longbench,ran2024novelqa,hamilton2025tldm}.
Screenplay resources emphasize analysis, summarization, or conditioned
generation
\citep{gorinski2015movie,papalampidi2020screenplay,saxena2024moviesum,chen2024hollmwood}.
These tasks establish the difficulty of long narrative context, but generally
evaluate the completed document through one output interface. STAGE aligns
three interfaces to the same temporal backbone. Its QA comparison also
contrasts flat retrieval
\citep{robertson2009bm25,lewis2020rag,izacard2021fid}, hierarchical access such
as PageIndex \citep{zhang2025pageindex}, and graph-informed access
\citep{edge2024fromlocal,guo2025graphragfi,yang2025eventrag,gantt2024event,luo2025etrqa}.

\textbf{Character role-playing and memory.}
Persona-conditioned dialogue, fictional role adoption, character profiling,
and long-term agent memory are studied by Persona-Chat, CharacterGPT,
PersonaEval, TVShowGuess, PersoNet, and LoCoMo
\citep{zhang2018personalizing,li2023roleplay,park2025charactergpt,
mazumder2024personaeval,sang2022tvshowguess,yu2023personet,
maharana-etal-2024-evaluating}. Most systems condition behavior on a static
profile, a known role, or dialogue history. ICRP fixes a screenplay
checkpoint, so persona expression must remain compatible with both evolving
state and the information available at that point.

\textbf{LLM-assisted construction and evaluation.}
Open-ended benchmark construction and scoring increasingly use LLMs under
explicit quality controls
\citep{li-etal-2025-autobencher,bai-etal-2023-lmexaminer,
gu-etal-2025-structext,zheng2023judging}. Reference-conditioned and factuality
methods reduce unconstrained preference judgments by grounding decisions in
claims or evidence
\citep{badshah-sajjad-2025-reference,min2023factscore,liu2025verifact,
thorne2018fever,fan2020eliciting}. STAGE follows this evidence-first design:
reference answers, source scenes, contradiction targets, and knowledge-boundary
scaffolds are retained for evaluation, and judge reliability is calibrated
against blinded human annotations.

\begin{figure*}[t]
    \centering
    \includegraphics[width=\textwidth]{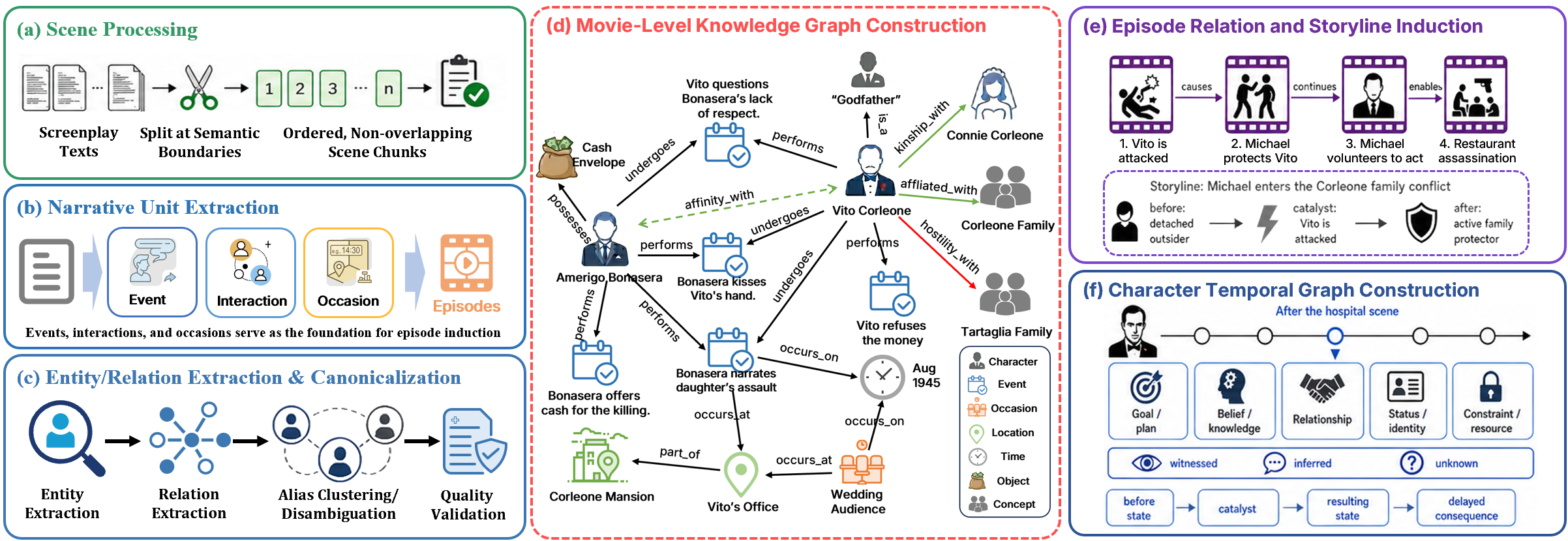}
    \caption{
      Six-stage construction of the STAGE narrative backbone, illustrated with
      \textit{The Godfather}: (a) scene processing; (b) narrative-unit
      extraction; (c) entity and relation canonicalization; (d) movie-level KG
      construction; (e) Episode-relation and Storyline induction; and (f)
      character temporal-graph construction. Every retained asset preserves its
      supporting screenplay scenes.
    }
    \label{fig:backbone_construction}
\end{figure*}

\section{The STAGE Framework}
\label{sec:dataset}

\subsection{Dataset}

\subsubsection{Data collection and statistics}
STAGE comprises 151 feature-length movie screenplays (109 English and 42
Chinese) collected from publicly accessible screenplay repositories and script
documents. We retain scripts with recoverable scene boundaries, dialogue, and
action descriptions, excluding summaries, transcripts, and fan-written
adaptations. After normalizing scene structure and OCR artifacts, we create the
primary original-identity corpus. The public release also includes
pseudonymized counterparts that differ only in approved work-specific
entity forms; scene order, content structure, and task semantics remain
unchanged. The scripts average about 30K words and 151 scenes;
detailed distributions by language, genre, release year, length, and scene count
appear in Appendix~\ref{sec:dataset_details}.

\subsubsection{Narrative backbone construction}
The STAGE narrative backbone is a provenance-linked, multilevel representation
that links screenplay evidence to movie-level narrative structure and to each
character's state at a given story point. It adapts the event--scene--plotline
progression in computational narratology \citep{piper-etal-2021-narrative};
Figure~\ref{fig:backbone_construction} summarizes six provenance-preserving
construction stages.

The first three stages normalize ordered scenes, extract source-bearing Events,
Interactions, and Occasions, and canonicalize entities and typed relations
(Figure~\ref{fig:backbone_construction}a--c). Ordered scenes are segmented at
semantic boundaries with complete source coverage. An \emph{Event} records a
bounded action or change, an \emph{Interaction} a process involving multiple
characters, and an \emph{Occasion} a reusable activity or situation. Each unit
retains its supporting text spans, while aliases and typed relations are
canonicalized in a shared entity registry.

The next two stages group scene-local units into Episodes and connect Episodes
across scenes as Storylines (Figure~\ref{fig:backbone_construction}d--e). Each
Event or Interaction belongs to a primary Episode that captures its setup,
development, and outcome. Storylines then connect these Episodes through an
explicit prior state, catalyst, resulting state, and scene provenance.

The final stage derives a temporal graph for each recurring character, with
ordered checkpoints, state and development ledgers, epistemic access, and
persona evidence (Figure~\ref{fig:backbone_construction}f). States cover goals,
knowledge, relationships, identity, and constraints. Developments connect a
prior state to its catalyst, resulting state, and possible delayed consequence;
epistemic links distinguish what the character witnessed, was told, inferred,
or does not know. Only information available to a character by a checkpoint may
enter its state or knowledge.

Episodes and Storylines provide evidence-linked paths for Task~II, while the
temporal graph supplies the checkpoint-valid states and knowledge boundaries
used by Tasks~I and III. These structures remain construction- and
evaluator-side. Appendices~\ref{app:kg_extraction}
and~\ref{app:release_human_audit} and Table~\ref{tab:backbone_task_support}
document the complete schemas, review, validation, and task mapping.

In a five-screenplay pilot, backbone guidance increased mean evidence span from
9.2\% to 46.5\% of a screenplay and audited reasoning paths from 1.5 to 3.6
logical hops. It also raised cross-scene evidence use from 26.0\% to 84.0\% and
grounded accuracy from 62.5\% to 94.2\%, while reducing narrative-logic
hallucinations from 36.0\% to 5.8\% (Appendix
Table~\ref{tab:backbone_necessity_pilot}). These results establish the
backbone's construction benefit within the pilot scope.

\subsubsection{Paired work-identity controls}
\label{sec:work_identity_construction}

Widely known films may appear in model pretraining data, allowing models to use
work-specific identity cues. STAGE pairs each original instance with a
title-blind pseudonymized counterpart derived from the frozen entity registry.
Reviewed one-to-one mappings propagate through screenplays, task inputs,
references, and evaluator evidence while preserving scene order, checkpoints,
states, evidence, and answer semantics. Paired differences measure sensitivity
to explicit identity cues. Distinctive narrative content may still reveal the
underlying work
\citep{yan-etal-2022-entity-renaming,yao-etal-2024-cross-language-contamination}.
Appendix~\ref{app:work_identity_construction} specifies the mapping, propagation,
and quality-control procedure.

\subsection{Tasks}
STAGE operationalizes evolving-story understanding as three complementary
capabilities: tracking character development, reasoning about narrative
evolution, and acting from a checkpoint-valid character state. Internal graph,
state, and provenance identifiers remain evaluator-side throughout all tasks.
The release contains 434 roles, 2{,}925 Task~I checkpoints and 7{,}912 states,
5{,}010 Task~II questions, and 5{,}425 single-turn plus 866 multi-turn Task~III
instances; Table~\ref{tab:overall_stats} gives the language breakdown.

\begin{table}[t]
\stageTableFont
\centering
\caption{Overall statistics of STAGE evaluation assets.}
\label{tab:overall_stats}
\setlength{\tabcolsep}{2.1pt}
\renewcommand{\arraystretch}{1.0}
\begin{tabular*}{\columnwidth}{@{\extracolsep{\fill}}lrrrrrr@{}}
\toprule
\multirow{2}{*}{\textbf{Language}} &
\multirow{2}{*}{\textbf{Roles}} &
\multicolumn{2}{c}{\textbf{Task I}} &
\multicolumn{1}{c}{\textbf{Task II}} &
\multicolumn{2}{c}{\textbf{Task III}} \\
\cmidrule(lr){3-4} \cmidrule(lr){5-5} \cmidrule(lr){6-7}
& &
\makecell{\textbf{Check-}\\\textbf{points}} &
\textbf{States} &
\textbf{QA} &
\makecell{\textbf{Single-}\\\textbf{turn}} &
\makecell{\textbf{Multi-}\\\textbf{turn}} \\
\midrule
English & 330 & 2,235 & 6,070 & 3,583 & 4,288 & 628 \\
Chinese & 104 & 690 & 1,842 & 1,427 & 1,137 & 238 \\
\midrule
\textbf{Total} & \textbf{434} & \textbf{2,925} & \textbf{7,912} & \textbf{5,010} & \textbf{5,425} & \textbf{866} \\
\bottomrule
\end{tabular*}
\end{table}

The three tasks share evidence, checkpoints, and temporal boundaries by
construction. Task~I samples successive slices of the character temporal graph and
asks models to update the state that holds at each checkpoint. Task~II follows
evidence-linked Episode and Storyline paths to ask what changed, when it became
effective, and which consequences persisted across scenes. Task~III converts a
checkpoint slice into model-visible persona, knowledge, relationship, and
memory context, then tests whether those constraints are used in generation.
This common provenance supports controlled comparison of state-maintenance,
evidence-integration, and grounded-generation failures across tasks.

STAGE is an evaluation-only benchmark. Tasks~I--III are released as fixed
evaluation sets without training, development, or test splits. The
development/test partition in
Appendix~\ref{app:entity_resolution_validation} belongs exclusively to the
separate 20-film validation of construction-time entity resolution.

\subsubsection{Task I: Character Development Tracking}
\label{sec:task1_structuring}

Given a focal character's preceding state and observations from the interval
between adjacent checkpoints, Task~I predicts the current state and newly
effective developments across goals, knowledge, relationships, identity, and
constraints. \textbf{Reference-State Update (RSU)} supplies the reviewed prior
state to isolate local updating; \textbf{Autoregressive State Update (ASU)}
supplies the model's prior prediction to expose accumulated error. Inputs,
targets, and prompts otherwise remain fixed, with change, no-change,
inaccessible-event, delayed-consequence, and final controls. Full eligibility,
construction, and a worked trajectory appear in
Appendices~\ref{app:task1_construction} and \ref{app:task1_walkthrough}.

\subsubsection{Task II: Cross-Scene Narrative Evolution Reasoning}
\label{sec:stageqa}

Task~II contains 5{,}010 questions requiring source-grounded reasoning across
scenes. Its six types cover Scene Grounding, Character Understanding,
Causal-Motivational Reasoning, Temporal Reasoning, Narrative Progression, and
Role-Relation Continuity. Each item has a reference answer and traceable support
for a required cross-scene path, following the anti-shortcut motivation of
multi-hop QA
\citep{yang-etal-2018-hotpotqa,ho-etal-2020-2wikimultihopqa,trivedi-etal-2022-musique}
(Appendices~\ref{app:qa_construction} and \ref{app:task2_walkthrough}).

\subsubsection{Task III: In-Script Character Role-Playing}
\label{sec:stageicrp}

Task~III anchors a focal character and audience request at a screenplay
checkpoint. A frozen context pack supplies visible persona evidence, dialogue
style, acquired memories, relationships, and the current user turn. Responses
must remain natural, in character, and bounded by checkpoint-available
knowledge. Single-turn interaction is primary; paired checkpoints and an
auxiliary multi-turn protocol test appropriate change, stability, knowledge
acquisition, and drift. Construction and a paired example appear in
Appendices~\ref{app:icrp_persona} and \ref{app:task3_walkthrough}.

\subsection{Evaluation}
\label{sec:evaluation}

STAGE separates local semantic judgments from deterministic metric aggregation.
The judge is independent of the evaluated model and calibrated against blinded
human annotations. Confidence intervals resample movies as dependency clusters.

\textbf{Task I.} We report \emph{Current-State Coverage}, \emph{Development
Coverage}, \emph{State Contradiction Rate}, and \emph{Development Contradiction
Rate}. Current-State and Development Coverage measure how completely a
prediction recovers, respectively, the state valid at the checkpoint and the
developments that became effective since the preceding checkpoint; the two
contradiction rates measure the share of predicted claims in each pool that
conflict with screenplay evidence or the checkpoint-valid state. Reference
claims receive full, partial, missing, or contradictory labels;
full and partial coverage score 1 and 0.5. ASU--RSU Development Coverage
measures recursive degradation, and headline scores use movie-macro aggregation
(Appendix~\ref{app:task1_eval}).

\textbf{Task II.} Each generated answer receives a binary correctness judgment
against the reference and screenplay evidence. We report Average Accuracy over
five generations and Pass@5; citation support remains a separate diagnostic
(Appendix~\ref{app:stageqa_eval}).

\textbf{Task III.} Each response receives four independent 1--5 scores:
Character Fidelity, Memory Faithfulness, Boundary Compliance, and Response
Naturalness. Typed checkpoint pairs test expected change, stability, knowledge
acquisition, and relationship change; we report the dimensions separately
(Appendix~\ref{app:stageicrp_eval}).

\section{Experimental Setup}
\label{sec:exp_setup}

\subsection{Research Questions}

\textbf{RQ1: How accurately can models track character development over time?}
We test integration of new evidence into current states and developments;
RSU--ASU differences isolate error accumulated through recursive state reuse.
Separate coverage and contradiction metrics distinguish omission from explicit
state conflict.

\textbf{RQ2: How reliably can models reason about narrative evolution across
scenes?} We compare flat, hierarchical, and graph-based access on questions
about cross-scene change and persistence. Pass@5 measures whether a system can
produce at least one correct answer, while average accuracy measures stability
across repeated generations.

\textbf{RQ3: Can models act consistently from a checkpoint-bounded character
state?} We vary memory access and compare typed checkpoints to test natural,
in-character behavior under checkpoint-available information. Independent
dimensions distinguish persona fidelity, screenplay-memory use, epistemic
boundary compliance, and response quality.

\subsection{Models and Baselines}

We evaluate Qwen3 (8B, 30B-A3B, and 235B), Llama-3.1 (8B and 70B), GPT-5.5,
Gemini 3.1 Pro, and Claude Sonnet 4.6 with standardized prompts
\citep{yang2025qwen3,grattafiori2024llama3,openai2026gpt55,
google2026gemini31pro,anthropic2026sonnet46}. Parameter counts are reported
where disclosed by the provider or encoded in the model name; undisclosed
counts for proprietary models remain unreported. The open-weight
Qwen3 and Llama-3.1 checkpoints are served locally on the single-node cluster
described in Appendix~\ref{app:exp_details}, using FP8 or AWQ quantization where
required by GPU-memory constraints.

Task~I compares RSU and ASU under matched intervals and observations. Task~II
compares \textbf{Hybrid RAG}, which combines dense and lexical scene retrieval,
hierarchical \textbf{PageIndex}
\citep{zhang2025pageindex} and \textbf{A-RAG}
\citep{du2026arag}, and graph-based access through \textbf{GraphRAG}
\citep{edge2024fromlocal} and \textbf{LightRAG}
\citep{guo-etal-2025-lightrag}. The Kimi~2.6 full-screenplay reference
\citep{moonshot2026kimi26} uses a different access setting and therefore
appears in Appendix~\ref{app:kimi_full_context_results}. All retrieval systems
operate over the same screenplay release and reader models, isolating the
effects of memory organization and evidence selection.

Task~III fixes persona context and varies episodic memory: Persona Card Only,
all checkpoint-visible memories, or the top five memories from BM25 or vector
retrieval.

\subsection{Evaluation Protocol}

\textbf{DeepSeek-V4-Pro \citep{deepseek2026v4pro} serves as the judge for all
tasks}, independently of the
evaluated model; deterministic code converts its judgments into the metrics in
Section~\ref{sec:evaluation}. Judge calls use the official DeepSeek API,
separately from the local FP8 construction deployment described in
Appendix~\ref{app:exp_details}. On 1{,}000 stratified outputs, it agreed with
adjudicated human consensus in 90.3\% of cases
(Appendix~\ref{app:cross_task_judge_reliability}); the two blinded human raters
agreed in 91.7\% before adjudication. The calibration sample contains 200
Task~I state outputs, 500 Task~II answers, and 300 open-ended Task~III
responses, reflecting their different judgment structures. Judge--consensus
agreement is 90.2\%, 93.0\%, and 85.8\%, respectively, with Cohen's $\kappa$ of
0.82, 0.86, and 0.76. The lower Task~III agreement reflects its graded,
open-ended dimensions.

All systems use the frozen release with temperature $T=0.2$, top-$p=0.9$, and at
most 8{,}192 generated tokens. Task~I and Task~III use one generation per
instance; Task~II uses five. Prompts, model IDs, decoding settings, hashes,
failures, and denominators are logged, and confidence intervals resample movies
as dependency clusters.

The model-facing templates, decoding parameters, output budgets, and
task-specific memory conditions are fixed uniformly across evaluated models.
Prompts exclude benchmark items paired with their gold labels or reference
answers as in-context demonstrations. Worked examples in the appendix serve as
reader illustrations and are excluded from evaluation inputs.

\section{Results and Analysis}
\label{sec:results}

Results address tracking, reasoning, and generation on the original benchmark
view; paired pseudonymized runs audit sensitivity to explicit work identity.

\subsection{RQ1: Character Development Tracking}

\begin{table*}[htbp]
\centering
\stageTableFont
\setlength{\tabcolsep}{2.5pt}
\begin{tabular*}{\textwidth}{@{\extracolsep{\fill}}l cccc cccc@{}}
\toprule
\multirow{2}{*}{\textbf{Model}} &
\multicolumn{4}{c}{\textbf{Reference-State Update}} &
\multicolumn{4}{c}{\textbf{Autoregressive State Update}} \\
\cmidrule(lr){2-5}\cmidrule(lr){6-9}
& \makecell{\textbf{State}\\\textbf{Cov.}}
& \makecell{\textbf{Dev.}\\\textbf{Cov.}}
& \makecell{\textbf{State}\\\textbf{Contr.}}
& \makecell{\textbf{Dev.}\\\textbf{Contr.}}
& \makecell{\textbf{State}\\\textbf{Cov.}}
& \makecell{\textbf{Dev.}\\\textbf{Cov.}}
& \makecell{\textbf{State}\\\textbf{Contr.}}
& \makecell{\textbf{Dev.}\\\textbf{Contr.}} \\
\midrule
Qwen3-8B          & 48.20 & 42.50 & 2.60 & 15.10 & 36.50 & 31.00 & 3.80 & 19.40 \\
Qwen3-30B-A3B     & 61.80 & 65.40 & 1.25 & 8.10  & 56.20 & 60.10 & 1.40 & 8.85 \\
Qwen3-235B        & 66.46 & 72.11 & 0.97 & 6.56  & 63.39 & 70.38 & 0.67 & 6.66 \\
Llama-3.1-8B      & 42.10 & 36.80 & 3.10 & 18.50 & 29.40 & 23.20 & 4.90 & 23.80 \\
Llama-3.1-70B     & 54.60 & 52.30 & 1.85 & 11.20 & 47.10 & 44.80 & 2.30 & 13.50 \\
GPT-5.5           & \textbf{74.50} & \textbf{78.80} & 0.42 & \textbf{4.10}
                  & \textbf{72.10} & \textbf{76.50} & 0.45 & \textbf{4.35} \\
Gemini 3.1 Pro    & 60.10 & 62.80 & 1.10 & 7.90  & 55.30 & 57.60 & 1.35 & 8.40 \\
Claude Sonnet 4.6 & 68.20 & 73.50 & \textbf{0.35} & 4.80
                  & 65.40 & 71.00 & \textbf{0.38} & 5.10 \\
\bottomrule
\end{tabular*}
\caption{
Task~I results under Reference-State Update (RSU) and Autoregressive State
Update (ASU) on the original, non-pseudonymized release. All values are movie-macro
percentages. Bold marks the best value within each update setting; Contr.
denotes Contradiction.
}
\label{tab:task1_results}
\end{table*}

\textbf{GPT-5.5} leads both coverage measures under RSU and ASU, while
\textbf{Claude Sonnet 4.6} has the lowest State Contradiction
(Table~\ref{tab:task1_results}). GPT-5.5 reaches 74.50/78.80 Current-State/
Development Coverage under RSU and 72.10/76.50 under ASU; even these maxima
leave substantial content uncovered, and low contradiction does not imply
complete tracking. Coverage declines from RSU to ASU for every model. With
scale, the Development Coverage loss narrows from 11.50 to 1.73 points for Qwen
and from 13.60 to 7.50 for Llama, while smaller models produce more
contradictions. The protocol gap therefore reflects lost ASU coverage more than
additional incompatible claims. Recursive failure appears primarily as silent
forgetting: scaling mitigates, but does not remove, the difficulty of
maintaining a trajectory across updates.

Pseudonymization shifts RSU and ASU coverage in opposite directions, but no
coverage or contradiction change is reliable (all 95\% intervals include zero;
Appendix Table~\ref{tab:task1_identity_effects}). Memorized identity effects
remain unresolved, and between-model differences are descriptive because
Table~\ref{tab:task1_results} reports point estimates.

\subsection{RQ2: Cross-Scene Narrative Evolution Reasoning}

\begin{table*}[htbp]
\centering
\stageTableFont
\setlength{\tabcolsep}{2.5pt}
\begin{tabular*}{\textwidth}{@{\extracolsep{\fill}}l cc cc cc cc cc@{}}
\toprule
\multirow{2}{*}{\textbf{Model}} &
\multicolumn{2}{c}{\textbf{Hybrid RAG}} &
\multicolumn{2}{c}{\textbf{PageIndex}} &
\multicolumn{2}{c}{\textbf{A-RAG}} &
\multicolumn{2}{c}{\textbf{GraphRAG}} &
\multicolumn{2}{c}{\textbf{LightRAG}} \\
\cmidrule(lr){2-3}\cmidrule(lr){4-5}\cmidrule(lr){6-7}\cmidrule(lr){8-9}\cmidrule(lr){10-11}
& \textbf{Pass@5} & \textbf{Avg. Acc.}
& \textbf{Pass@5} & \textbf{Avg. Acc.}
& \textbf{Pass@5} & \textbf{Avg. Acc.}
& \textbf{Pass@5} & \textbf{Avg. Acc.}
& \textbf{Pass@5} & \textbf{Avg. Acc.} \\
\midrule
Qwen3-8B          & 46.1 & 42.4 & 34.9 & 18.0 & 27.7 & 23.6 & 31.8 & 16.7 & 42.9 & 36.6 \\
Qwen3-30B-A3B     & 55.3 & 51.5 & 34.2 & 33.7 & 37.2 & 30.2 & 41.2 & 29.8 & 54.2 & 38.8 \\
Qwen3-235B        & 63.7 & 60.8 & 38.6 & 34.3 & 48.6 & 36.0 & 47.1 & 33.5 & 60.9 & 43.9 \\
Llama-3.1-8B      & 41.3 & 38.1 & 25.2 & 16.8 & 25.7 & 21.8 & 29.6 & 19.8 & 41.5 & 33.9 \\
Llama-3.1-70B     & 55.8 & 51.6 & 35.8 & 30.2 & 42.4 & 33.9 & 43.3 & 30.1 & 52.0 & 38.7 \\
GPT-5.5           & 65.0 & \textbf{62.5} & 40.6 & 35.6 & \textbf{53.7} & \textbf{42.1} & \textbf{51.7} & 39.2 & \textbf{61.6} & \textbf{47.9} \\
Gemini 3.1 Pro    & 63.9 & 61.2 & 40.6 & 35.6 & 51.4 & 40.6 & 51.5 & \textbf{40.6} & 60.2 & 46.5 \\
Claude Sonnet 4.6 & \textbf{65.1} & 61.6 & \textbf{44.1} & \textbf{40.0} & 52.4 & 41.2 & 50.3 & 39.3 & 60.1 & 46.4 \\
\bottomrule
\end{tabular*}
\caption{
Cross-Scene Narrative Evolution Reasoning results on the full, original Task~II
question-answering release. Pass@5 counts a question as
correct if any of five generations is judged correct; Average Accuracy averages
correctness across the five generations.
}
\label{tab:qa_results}
\end{table*}

\textbf{Hybrid RAG} performs best overall, but the strongest Pass@5 and average
accuracy reach only 65.1 and 62.5 (Table~\ref{tab:qa_results}). The oracle union
reaches 78.6\%, and Pass@5--accuracy gaps indicate inconsistent evidence reuse.
The oracle gain shows that retrieval systems recover complementary evidence,
yet no reader model converts it into consistently correct answers across
generations. Among alternative structures,
\textbf{LightRAG} \citep{guo-etal-2025-lightrag} outperforms \textbf{GraphRAG}
for every model, while \textbf{A-RAG} \citep{du2026arag} generally exceeds
PageIndex on larger models.

Removing screenplay context lowers accuracy by 41.7--51.6 points, whereas
pseudonymization has small, uncertain effects (Appendix
Tables~\ref{tab:task2_identity_audit} and
\ref{tab:task2_closed_book_identity}). Evidence access thus matters more than
explicit identity cues, although semantic content may still reveal the work.
Complete gold-scene coverage improves accuracy by 24.0--27.0 points, yet
27.1--29.6\% of answers remain incorrect and Temporal Reasoning reaches only
44.9--48.9\% (Appendix Table~\ref{tab:task2_retrieval_coverage}). Retrieval
coverage is therefore necessary but insufficient for ordered-state reasoning:
models still fail to integrate available scenes into the required temporal and
causal sequence. The category breakdown also shows retrieval specialization:
on Qwen3-235B, LightRAG leads
Character Understanding, Role-Relation Continuity, and Temporal Reasoning,
whereas Hybrid RAG leads Scene Grounding, Causal-Motivational Reasoning, and
Narrative Progression (Appendix Table~\ref{tab:qa_category_breakdown}). No
structure dominates all reasoning types; representative temporal and causal
failures appear in Table~\ref{tab:task2_failure_cases}.

\newcommand{\taskIIIFullSingleTurnTable}{%
\begin{table*}[!t]
\centering
\scriptsize
\setlength{\tabcolsep}{3.0pt}
\renewcommand{\arraystretch}{1.05}
\resizebox{\textwidth}{!}{%
\begin{tabular}{l*{16}{c}}
\toprule
\textbf{Metric}
& \multicolumn{4}{c}{\textbf{Qwen3-8B}}
& \multicolumn{4}{c}{\textbf{Qwen3-30B-A3B}}
& \multicolumn{4}{c}{\textbf{Qwen3-235B}}
& \multicolumn{4}{c}{\textbf{Llama-3.1-8B}} \\
\cmidrule(lr){2-5}\cmidrule(lr){6-9}\cmidrule(lr){10-13}\cmidrule(lr){14-17}
& PC & FM & BM25 & Vect
& PC & FM & BM25 & Vect
& PC & FM & BM25 & Vect
& PC & FM & BM25 & Vect \\
\midrule
Fidelity
& 4.33 & 4.51 & \textbf{4.69} & 4.68
& 4.42 & 4.62 & \textbf{4.78} & 4.77
& 4.52 & 4.72 & \textbf{4.83} & 4.83
& 3.98 & 4.12 & \textbf{4.35} & 4.35 \\
Natural.
& 4.72 & 4.79 & \textbf{4.82} & 4.82
& 4.82 & 4.89 & \textbf{4.92} & 4.92
& 4.84 & 4.93 & \textbf{4.95} & 4.95
& 4.65 & 4.72 & \textbf{4.77} & 4.76 \\
Boundary
& 3.86 & 4.25 & \textbf{4.38} & 4.38
& 4.26 & 4.55 & \textbf{4.68} & 4.68
& 4.63 & 4.79 & \textbf{4.89} & 4.88
& 3.69 & 4.08 & \textbf{4.21} & 4.21 \\
Mem. Faith
& 1.58 & \textbf{3.85} & 3.45 & 3.37
& 1.68 & \textbf{4.32} & 3.87 & 3.78
& 1.76 & \textbf{4.69} & 4.19 & 4.10
& 1.53 & \textbf{3.62} & 3.24 & 3.17 \\
\midrule
\textbf{Metric}
& \multicolumn{4}{c}{\textbf{Llama-3.1-70B}}
& \multicolumn{4}{c}{\textbf{GPT-5.5}}
& \multicolumn{4}{c}{\textbf{Gemini 3.1 Pro}}
& \multicolumn{4}{c}{\textbf{Claude Sonnet 4.6}} \\
\cmidrule(lr){2-5}\cmidrule(lr){6-9}\cmidrule(lr){10-13}\cmidrule(lr){14-17}
& PC & FM & BM25 & Vect
& PC & FM & BM25 & Vect
& PC & FM & BM25 & Vect
& PC & FM & BM25 & Vect \\
\midrule
Fidelity
& 4.32 & 4.68 & \textbf{4.85} & 4.85
& 4.76 & 4.88 & \textbf{4.97} & 4.96
& 4.69 & 4.81 & \textbf{4.93} & 4.92
& 4.71 & 4.83 & \textbf{4.93} & 4.92 \\
Natural.
& 4.75 & 4.83 & \textbf{4.87} & 4.86
& 4.91 & 4.98 & \textbf{4.99} & 4.99
& 4.88 & 4.95 & \textbf{4.97} & 4.96
& 4.90 & 4.96 & \textbf{4.99} & 4.98 \\
Boundary
& 4.13 & 4.44 & \textbf{4.57} & 4.57
& 4.82 & 4.93 & \textbf{4.98} & 4.98
& 4.69 & 4.82 & \textbf{4.91} & 4.90
& 4.76 & 4.88 & \textbf{4.95} & 4.95 \\
Mem. Faith
& 1.65 & \textbf{4.18} & 3.74 & 3.66
& 1.82 & \textbf{4.96} & 4.44 & 4.34
& 1.79 & \textbf{4.81} & 4.31 & 4.21
& 1.80 & \textbf{4.88} & 4.37 & 4.27 \\
\bottomrule
\end{tabular}
}
\caption{Complete Task~III single-turn results across all eight models and four
memory settings on the original release (1--5 scale). Bold marks the best
setting per model and metric. PC: Persona Card Only; FM: Full Memory Context;
Vect: vector retrieval.}
\label{tab:icrp_results}
\end{table*}
}

\newcommand{\taskIIIFullMultiTurnTable}{%
\begin{table*}[htbp]
\centering
\scriptsize
\setlength{\tabcolsep}{3.0pt}
\renewcommand{\arraystretch}{1.05}
\resizebox{\textwidth}{!}{%
\begin{tabular}{l*{16}{c}}
\toprule
\textbf{Metric}
& \multicolumn{4}{c}{\textbf{Qwen3-8B}}
& \multicolumn{4}{c}{\textbf{Qwen3-30B-A3B}}
& \multicolumn{4}{c}{\textbf{Qwen3-235B}}
& \multicolumn{4}{c}{\textbf{Llama-3.1-8B}} \\
\cmidrule(lr){2-5}\cmidrule(lr){6-9}\cmidrule(lr){10-13}\cmidrule(lr){14-17}
& PC & FM & BM25 & Vect
& PC & FM & BM25 & Vect
& PC & FM & BM25 & Vect
& PC & FM & BM25 & Vect \\
\midrule
Fidelity
& 3.74 & 3.42 & 3.68 & 3.90
& 4.89 & 4.81 & 4.90 & 4.85
& 4.99 & 4.96 & 4.96 & 4.94
& 3.69 & 3.24 & 3.52 & 3.71 \\
Natural.
& 3.51 & 3.39 & 3.41 & 3.52
& 4.53 & 4.40 & 4.29 & 4.42
& 4.74 & 4.76 & 4.70 & 4.76
& 3.44 & 3.24 & 3.27 & 3.35 \\
Boundary
& 3.61 & 3.68 & 3.65 & 3.72
& 4.02 & 3.92 & 3.88 & 3.97
& 4.62 & 4.53 & 4.43 & 4.47
& 3.56 & 3.53 & 3.56 & 3.50 \\
Mem. Faith
& 1.52 & 2.46 & 2.19 & 2.21
& 1.35 & 2.51 & 2.13 & 2.18
& 1.58 & 3.23 & 2.64 & 2.55
& 1.46 & 2.32 & 2.10 & 2.03 \\
Cross-turn
& 4.12 & 4.15 & 4.18 & 4.21
& 4.97 & 4.97 & 4.94 & 4.94
& 4.99 & 4.99 & 4.99 & 4.99
& 4.31 & 4.33 & 4.27 & 4.35 \\
Follow-up
& 3.62 & 3.89 & 3.72 & 3.77
& 3.93 & 4.16 & 3.98 & 3.95
& 4.04 & 4.15 & 4.05 & 4.05
& 3.55 & 3.73 & 3.60 & 3.56 \\
\midrule
\textbf{Metric}
& \multicolumn{4}{c}{\textbf{Llama-3.1-70B}}
& \multicolumn{4}{c}{\textbf{GPT-5.5}}
& \multicolumn{4}{c}{\textbf{Gemini 3.1 Pro}}
& \multicolumn{4}{c}{\textbf{Claude Sonnet 4.6}} \\
\cmidrule(lr){2-5}\cmidrule(lr){6-9}\cmidrule(lr){10-13}\cmidrule(lr){14-17}
& PC & FM & BM25 & Vect
& PC & FM & BM25 & Vect
& PC & FM & BM25 & Vect
& PC & FM & BM25 & Vect \\
\midrule
Fidelity
& 3.86 & 3.31 & 3.70 & 3.70
& 4.97 & 4.96 & 4.98 & 4.98
& 4.99 & 4.53 & 4.96 & 5.00
& 4.98 & 4.99 & 4.99 & 4.99 \\
Natural.
& 3.89 & 3.56 & 3.71 & 3.70
& 4.97 & 4.98 & 4.96 & 4.97
& 4.24 & 3.55 & 4.20 & 4.18
& 4.99 & 4.54 & 4.68 & 4.81 \\
Boundary
& 4.00 & 3.79 & 3.79 & 3.88
& 4.91 & 4.81 & 4.86 & 4.89
& 4.69 & 3.60 & 4.07 & 4.18
& 4.95 & 4.51 & 4.83 & 4.68 \\
Mem. Faith
& 1.58 & 2.72 & 2.39 & 2.39
& 1.76 & 3.72 & 3.54 & 3.41
& 1.65 & 3.52 & 3.28 & 3.12
& 1.70 & 4.18 & 3.31 & 3.48 \\
Cross-turn
& 4.80 & 4.78 & 4.73 & 4.73
& 4.99 & 4.99 & 4.99 & 4.99
& 4.99 & 4.82 & 4.91 & 4.99
& 4.98 & 4.99 & 4.98 & 4.99 \\
Follow-up
& 3.72 & 3.84 & 3.79 & 3.74
& 4.21 & 4.45 & 4.18 & 4.09
& 3.69 & 3.48 & 3.63 & 3.73
& 4.10 & 3.98 & 4.31 & 3.85 \\
\bottomrule
\end{tabular}
}
\caption{
Complete Task~III multi-turn results across all eight models and four
memory-access settings on the original release (1--5 scale). Cross-turn denotes
Cross-turn Consistency; Follow-up denotes Follow-up Compatibility. PC: Persona
Card Only; FM: Full Memory Context; Vect: vector retrieval.
}
\label{tab:icrp_multiturn_full}
\end{table*}
}

\taskIIIFullSingleTurnTable

\subsection{RQ3: Grounded Persona Consistency in In-Script Character Role-Playing}
Task~III separates surface persona imitation from screenplay-grounded
role-playing. Persona-only inputs remain fluent and in character, yet Memory
Faithfulness is only 1.53--1.82; full memory raises it to 3.62--4.96 for every
model (Table~\ref{tab:icrp_results}). BM25 instead leads Fidelity, Naturalness,
and Boundary Compliance, with vector retrieval close behind. Fluent
characterization and accurate use of screenplay memory are therefore distinct
capabilities. BM25 concentrates generation on a small set of locally relevant
memories, supporting style and situation fidelity, but its top-five budget can
omit facts required for strict grounding; full memory supplies those facts
while introducing more competing episodes. Selective retrieval therefore
favors persona expression, whereas complete checkpoint-valid context favors
narrative recall. The consistency of this trade-off across models means that no
single access strategy dominates all four objectives; separate dimension-wise
scores prevent either behavior from being hidden by an aggregate.

The multi-turn test retains the full-memory advantage but shows model-specific
drift (Appendix Table~\ref{tab:icrp_multiturn_full}). Support diagnostics further
distinguish retrieving a relevant memory from using it correctly
(Appendix~\ref{app:roleplay_support_access}), confirming that access alone does
not ensure grounding. Identity audits leave Memory Faithfulness, Naturalness,
and multi-turn continuity broadly stable, with small Fidelity and Boundary
Compliance shifts (Appendix Tables~\ref{tab:task3_identity_effects} and
\ref{tab:task3_multiturn_identity}). These effects remain diagnostic rather than
a general contamination estimate. Representative failures expose the remaining
gap: dialogue can sound natural while inventing unsupported premises or ignoring
checkpoint-valid interaction context (Appendix
Table~\ref{tab:task3_failure_cases}).

\FloatBarrier

\section{Conclusion}
\label{sec:conclusion}

STAGE evaluates full-screenplay understanding through three aligned capabilities:
tracking character development, reasoning across scenes, and role-playing within
checkpoint-valid state and knowledge boundaries. Across 151 English and Chinese
films, the tasks expose complementary failures in maintaining, integrating, and
applying evolving narrative state. Recursive updates lose prior developments,
cross-scene QA remains difficult even with complete annotated evidence, and
persona expression can diverge from screenplay faithfulness.
These findings show that context length, retrieval coverage, and fluent
generation are incomplete proxies for narrative understanding. Together, the
shared provenance-linked backbone supports diagnosis of state updating,
evidence integration, memory, epistemic control, and identity effects.

\clearpage

\section{Limitations}
\label{sec:limitations}

STAGE has three main limitations.

\paragraph{Construction quality and cost.}
The backbone and task assets use a hybrid human--LLM pipeline, so residual
annotation errors may remain, particularly in Tasks~I and III where audits are
representative. Task~II receives exhaustive review, while risk-based checks,
source-level provenance, and deterministic validation make all retained items
traceable. These controls prioritize release quality, although construction
cost still limits throughput.

\paragraph{Context formulation.}
Performance depends partly on how evidence is organized and accessed. Task~I
uses fixed checkpoint intervals, Task~II compares retrieval systems, and
Task~III varies checkpoint-visible memory. STAGE standardizes prompts and access
conditions within each comparison and reports retrieval and memory ablations.
The resulting claims therefore concern model behavior under explicit,
documented context conditions.

\paragraph{Coverage and identity.}
The corpus covers English and Chinese movie screenplays, with fewer Chinese
films because available sources are less centralized and less consistently
formatted. The bilingual release broadens language coverage, while the source
index and reconstruction tools provide a basis for expansion. The released
pseudonymized view reduces explicit title and entity cues, although distinctive
semantic content may still reveal a work. Paired differences thus quantify
sensitivity to work identity; estimating training-data contamination requires
separate evidence.

\section{Ethical Considerations}
\label{sec:data_usage_ethics}

STAGE follows the linked-source release design of
NarrativeQA~\citep{kovcisky2018narrativeqa}. The public package contains source
metadata and links, integrity information, an original-identity reconstruction
tool, STAGE-authored annotations, and evaluation code. Users obtain screenplay
text from its recorded source and materialize the original-identity view
locally. The public package also includes the pseudonymized control used for
the work-identity robustness results. The STAGE-authored assets are provided for non-commercial research with
source attribution and a contact and removal procedure; ownership and
commercial rights remain with the source-work holders.

Preprocessing removes source front matter and incidental contact information,
and derived assets are screened before release. All annotation was conducted
voluntarily by adult coauthor domain experts. Because screenplays may contain
mature fictional content, the dataset card flags this possibility and
recommends appropriate access controls for research use.

LLMs assist structured annotation and evaluation, with source provenance,
human review, deterministic validation, and judge calibration providing quality
controls. STAGE is intended for controlled research evaluation of narrative
understanding. Its task-specific metrics are diagnostic signals, with broader
literary judgment beyond their scope.

\begingroup
\makeatletter
\let\stageoldthebibliography\thebibliography
\renewcommand{\thebibliography}[1]{%
  \stageoldthebibliography{#1}%
  \setlength{\itemsep}{0.2ex}%
  \setlength{\parsep}{0pt}%
}
\makeatother
\bibliography{acl_lualatex}
\endgroup

\appendix
\setcounter{table}{0}
\setcounter{figure}{0}
\renewcommand{\thetable}{A\arabic{table}}
\renewcommand{\thefigure}{A\arabic{figure}}
\renewcommand{\theHtable}{appendix.table.\arabic{table}}
\renewcommand{\theHfigure}{appendix.figure.\arabic{figure}}

\section{Dataset Details}
\label{sec:dataset_details}

\subsection{Data Collection, Annotation, and Release Policy}
\label{sec:data_collection}

STAGE is built from full-length movie screenplays collected from publicly
available sources. English scripts are primarily gathered from online screenplay
repositories indexed by IMDb~\cite{imdb}, while Chinese scripts are collected
from publicly accessible PDF or Word documents. These Chinese sources are often
substantially less standardized in structure, with variation in scene markers,
title versus numeric indexing, and the presentation of dialogue and action
lines. We exclude plot summaries, transcripts, and fan-written adaptations, and
retain only screenplay texts that preserve scene boundaries, dialogue, and
action descriptions.

\paragraph{Release package and governance.}
The public release contains a source index, retrieval and integrity metadata,
the local reconstruction tool, STAGE-authored annotations, and evaluation code.
Users obtain screenplay text from its recorded source, after which the tool
creates the scene-structured original-identity view while preserving names and
title cues. The work-identity analysis reported in this paper uses the released
pseudonymized counterparts; private identity mappings remain author-side.

Each source entry records available provenance, retrieval metadata, and
integrity information. Each locally processed screenplay also has validation
records. The package distinguishes source metadata, STAGE-authored
scene normalization and structural annotations, derived narrative and task
assets, and evaluation code. The STAGE-authored assets are released for
non-commercial research with source attribution and a documented contact and
removal procedure.

Source preprocessing removes front matter and incidental contact metadata,
including cover pages, headers, contact details, agency addresses, and
signatures, before downstream extraction. Derived entity records and QA assets
are additionally screened for residual contact information before release.

Seven bilingual coauthor-annotators provide human oversight, drawing on
professional expertise in screenwriting, directing, and narrative consulting.
LLM extraction and generation use fixed schemas and
prompts; annotators exhaustively review provisional Task~II items and audit
representative Task~I claims and Task~III interactions, with additional review
of low-confidence outputs, schema violations, and logical inconsistencies. The
same criteria are applied to English and Chinese screenplays; detailed audit
coverage appears in Appendix~\ref{app:release_human_audit}.

\subsection{Corpus and Backbone Statistics}
\label{sec:dataset_statistics}
\label{sec:kg_statistics}

\paragraph{Movie Genre Distribution.}
Each movie is annotated with up to three genre labels used only as descriptive
metadata. To reduce sparsity, low-frequency stylistic variants are merged into
 parent genres. Table~\ref{tab:genre_statistics} reports the aggregated
distribution across all genre annotation slots, so counts do not sum to the
number of movies.

\begin{table}[t]
    \centering
    \stageTableFont
    \setlength{\tabcolsep}{2pt}
    \renewcommand{\arraystretch}{1.05}
    \begin{tabular}{lrr@{\hspace{3pt}}lrr}
        \toprule
        Genre & Count & Prop. & Genre & Count & Prop. \\
        \midrule
        Drama & 78 & 17.0 & Horror & 13 & 2.8 \\
        Comedy & 59 & 12.9 & Family & 12 & 2.6 \\
        Crime & 42 & 9.2 & Sports & 11 & 2.4 \\
        Action & 38 & 8.3 & War & 8 & 1.7 \\
        Thriller & 38 & 8.3 & Social & 7 & 1.5 \\
        Romance & 33 & 7.2 & Mystery & 6 & 1.3 \\
        Adventure & 23 & 5.0 & Social Issues & 6 & 1.3 \\
        Fantasy & 21 & 4.6 & Music & 5 & 1.1 \\
        Science Fiction & 19 & 4.1 & Youth & 5 & 1.1 \\
        History & 15 & 3.3 & Others & 20 & 4.4 \\
        \bottomrule
    \end{tabular}
    \caption{Genre distribution in STAGE after taxonomy consolidation. Statistics are aggregated across all genre annotation slots (up to three per movie) over both English and Chinese films.}
    \label{tab:genre_statistics}
\end{table}

\begin{figure}[t]
    \centering
    \includegraphics[width=\columnwidth]{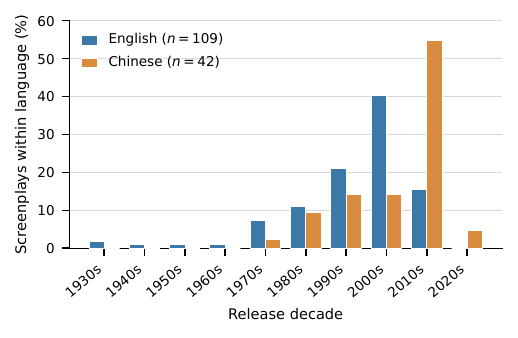}
    \caption{
    Release-decade distribution of STAGE screenplays. Bars show the percentage
    within each language subset to account for their different sizes. All 151
    screenplays have valid release-year metadata.
    }
    \label{fig:release_year_distribution}
\end{figure}

\paragraph{Release-Year Distribution.}
The 151 screenplays span releases from 1931 to 2023. The English subset
($n=109$) ranges from 1931 to 2015, with a median release year of 2001; the
Chinese subset ($n=42$) ranges from 1979 to 2023, with a median of 2011.5.
Figure~\ref{fig:release_year_distribution} reports decade-level proportions
within each language, rather than raw counts, because the two subsets differ in
size. These metadata characterize the corpus's temporal coverage; the paired
pseudonymization and closed-book analyses provide the direct tests of work-
identity effects.

\paragraph{Screenplay Length and Scene Count Distribution.}
We measure screenplay length as the total word count of scene titles,
subtitles, and contents in the locally materialized screenplay records, and structural
complexity as the number of scene records. Table~\ref{tab:length_statistics}
reports summary statistics by language; screenplay lengths range from 12{,}702
to 70{,}832 words and scene counts from 12 to 600.

\begin{figure*}[t]
    \centering
    \includegraphics[width=\textwidth]{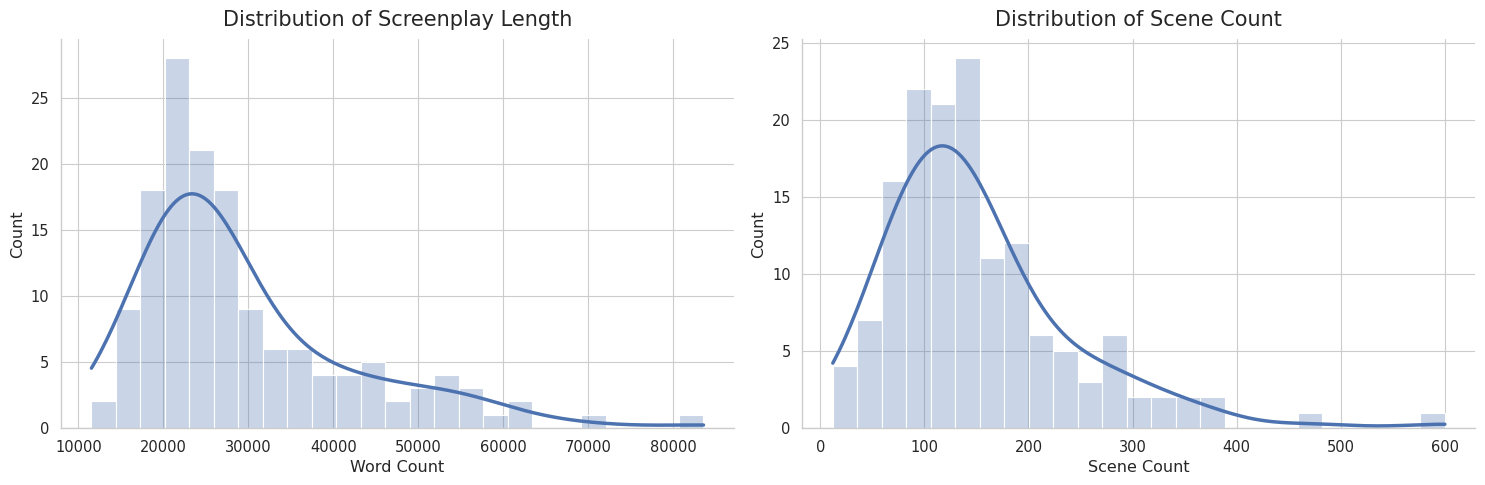}
    \caption{
    Distribution of screenplay lengths and scene counts in STAGE.
    The left histogram shows the distribution of screenplay word counts,
    while the right histogram shows the distribution of scene counts
    across all movies in the dataset.
    }
    \label{fig:length_scene_histogram}
\end{figure*}

\begin{table*}[t]
\centering
\stageTableFont
\setlength{\tabcolsep}{6pt}
\begin{tabular}{l r ccc ccc}
\toprule
& & \multicolumn{3}{c}{\textbf{Screenplay Length (words)}} &
\multicolumn{3}{c}{\textbf{Scene Count}} \\
\cmidrule(lr){3-5} \cmidrule(lr){6-8}
\textbf{Language} & \textbf{\#Movies} &
\textbf{Min} & \textbf{Mean} & \textbf{Max} &
\textbf{Min} & \textbf{Mean} & \textbf{Max} \\
\midrule
English  & 109 & 14{,}138 & 29{,}506 & 51{,}937 & 18  & 167 & 600 \\
Chinese  & 42  & 12{,}702 & 31{,}530 & 70{,}832 & 12 & 110 & 373 \\
\bottomrule
\end{tabular}
\caption{
Summary statistics of screenplay length (in words) and scene count for English
and Chinese films in STAGE.
}
\label{tab:length_statistics}
\end{table*}

\paragraph{Backbone Schema.}
The canonical KG uses a fixed seven-type entity registry
(Table~\ref{tab:kg_entity_schema}). Events and Interactions remain
evidence-bearing records outside the KG node registry and are later assigned to
Episodes. Occasions serve as reusable narrative contexts and may therefore be
canonical entities and relation endpoints. KG relations link
canonical entities under a fixed predicate registry and retain the validation
fields summarized in Table~\ref{tab:kg_relation_schema}.

\begin{table}[t]
\centering
\stageTableFont
\setlength{\tabcolsep}{5.5pt}
\begin{tabular}{p{1.8cm} p{4.9cm}}
\toprule
\textbf{Entity Type} & \textbf{Description} \\
\midrule
Character
& Concrete roles in the story, including human or anthropomorphized characters. \\

Occasion
& Concrete activities or situations that can organize multiple events, such as
meetings, confrontations, investigations, or ceremonies. \\

Location
& Physical, geographical, or virtual places and scenes where events occur. \\

TimePoint
& Specific points in time or time spans referenced in the narrative. \\

Object
& Key items, props, or devices that play an important role in the story. \\

Concept
& Abstract categories, ideas, species, or other non-organizational concepts. \\

Organization
& Institutions, factions, companies, agencies, or other organized groups that
can participate in narrative relations. \\
\bottomrule
\end{tabular}
\caption{
Entity types in the shared narrative backbone schema.
}
\label{tab:kg_entity_schema}
\end{table}

\begin{table*}[t]
\centering
\stageTableFont
\setlength{\tabcolsep}{4pt}
\renewcommand{\arraystretch}{1.15}
\begin{tabular}{>{\raggedright\arraybackslash}p{3.0cm} >{\raggedright\arraybackslash}p{10.0cm}}
\toprule
\textbf{Field} & \textbf{Constraint} \\
\midrule
Canonical endpoints
& Both endpoints must resolve to canonical entity identifiers; unresolved
surface names and Event/Interaction records cannot serve as KG endpoints. \\
Predicate
& The predicate must belong to the frozen registry and satisfy its domain and
range constraints. The taxonomy is not expanded in response to extraction
errors. \\
Qualifiers
& Polarity, modality, and temporal scope are stored explicitly when supported
by the screenplay. \\
Source evidence
& Every retained relation stores source-scene identifiers and original-text
evidence sufficient for later verification. \\
Deduplication key
& Relations with identical canonical endpoints, predicate, qualifiers, and
temporal scope are merged deterministically while preserving all evidence IDs. \\
\bottomrule
\end{tabular}
\caption{
Validation schema for canonical KG relations. The frozen predicate inventory
and domain--range registry are released with the construction configuration.
}
\label{tab:kg_relation_schema}
\end{table*}

\paragraph{Graph Size Statistics.}
Table~\ref{tab:kg_size_statistics} summarizes the canonical KG, induced
narrative units, Episodes, and Evolution Storylines per screenplay in the full
release.

\begin{table}[t]
\centering
\stageTableFont
\setlength{\tabcolsep}{2.5pt}
\resizebox{\columnwidth}{!}{%
\begin{tabular}{@{}l rr c rr@{}}
\toprule
& \multicolumn{2}{c}{\textbf{English ($N=109$)}}
& & \multicolumn{2}{c}{\textbf{Chinese ($N=42$)}} \\
\cmidrule{2-3} \cmidrule{5-6}
\textbf{Asset / Metric} & \textbf{Mean} & \textbf{Min--Max}
& & \textbf{Mean} & \textbf{Min--Max} \\
\midrule
\multicolumn{6}{l}{\textit{\textbf{Canonical KG Assets}}} \\
Canonical Entities & 346.1 & 222--431 & & 246.3 & 142--304 \\
Semantic Relations & 145.8 & 69--201  & & 71.0  & 48--95  \\
\midrule
\multicolumn{6}{l}{\textit{\textbf{Narrative Graph Assets}}} \\
\makecell[l]{Atomic Narrative\\Units$^\dagger$} & 483.0 & 285--578
& & 330.0 & 184--366 \\
Episodes & 175.3 & 81--230 & & 134.0 & 74--196 \\
\makecell[l]{Evolution\\Storylines} & 10.4 & 4--15 & & 7.7 & 3--12 \\
\bottomrule
\end{tabular}
}
\caption{
Summary statistics of canonical KG and narrative-graph assets per screenplay
across the full STAGE benchmark, reported after entity canonicalization and
graph validation. $^\dagger$Atomic narrative units aggregate evidence-bearing
Events, Interactions, and Occasions.
}
\label{tab:kg_size_statistics}
\end{table}

\subsection{Task Release Statistics}
\label{sec:task1_statistics}
\label{sec:qa_type_statistics}
\label{sec:icrp_statistics}
\paragraph{Character Development Tracking Assets.}
Task~I release statistics distinguish movies, focal-character trajectories,
and formal checkpoint instances. Movies define source documents and dependency
clusters; focal-character trajectories are the primary aggregation units; and
formal checkpoints are atomic scored observations. For each focal character,
Task~I contains an ordered checkpoint sequence, fixed adjacent-checkpoint
intervals, and evaluator-side rubrics grounded in state, development,
invariant, and future-negative evidence. Runtime processing produces the
interval memory exclusively for evaluation.
Checkpoint controls include actual changes, no-change intervals, inaccessible
events, delayed consequences, and final states. Their types remain explicit in
the released checkpoint metadata, supporting stratified analysis separately
from corpus, trajectory, and context-management counts.

\paragraph{Cross-Scene Evolution Question Taxonomy.}
Task~II uses the six-category narrative-reasoning taxonomy in
Table~\ref{tab:qa_type_statistics}. The categories and counts match the released
5{,}010-question package used in our experiments.

\begin{table*}[t]
    \centering
    \stageTableFont
    \renewcommand{\arraystretch}{1.25}
    \setlength{\tabcolsep}{4pt}
    \begin{tabular}{>{\raggedright\arraybackslash}p{4.0cm} c >{\raggedright\arraybackslash}p{7.8cm}}
        \toprule
        \textbf{Question Type} & \textbf{Count} & \textbf{Primary Reasoning Demand} \\
        \midrule
        \textbf{Scene Grounding} & 1{,}207 &
        Locate the relevant scene or scenes and identify the action, event,
        setting, or object grounded there. \\

        \textbf{Character Understanding} & 1{,}103 &
        Infer a character's local or evolving goals, intentions, knowledge,
        states, and decisions from screenplay evidence. \\

        \textbf{Causal-Motivational Reasoning} & 1{,}041 &
        Connect an action or development to its supported cause, trigger, or
        character motivation. \\

        \textbf{Temporal Reasoning} & 766 &
        Resolve event order, beginnings and endings, before--after relations,
        or the point at which a development occurs. \\

        \textbf{Narrative Progression} & 597 &
        Identify how turning points and earlier developments initiate, redirect,
        or shape a later plot trajectory. \\

        \textbf{Role-Relation Continuity} & 296 &
        Track social roles and relationships across scenes, including their
        persistence, escalation, weakening, or transformation. \\
        \bottomrule
    \end{tabular}
    \caption{
    Narrative-reasoning taxonomy and category counts for the released STAGE
    Task~II package (5{,}010 questions).
    }
\label{tab:qa_type_statistics}
\end{table*}

Reference answers are concise. English answers average 9.51 words (median 8;
range 1--50), while Chinese answers average 18.11 characters (median 15;
range 2--116). We report language-appropriate units because whitespace-based
word counts are not comparable for Chinese text.



\paragraph{In-Script Character Role-Playing Assets.}
Task~III contains checkpoint-bounded single-turn interactions and paired cases
for evolving, invariant, inaccessible, and post-disclosure conditions. Legacy
multi-turn interactions are retained as an auxiliary stress test; the released
counts are reported in Table~\ref{tab:overall_stats}.

\section{Benchmark Construction Details}
\label{app:construction}

This appendix details screenplay preprocessing, narrative-backbone construction
and validation, and task-specific asset generation.

\subsection{Data Preprocessing and Computational Environment}
\label{app:screenplay_processing}

Because screenplay formatting is heterogeneous, preprocessing combines
rule-based parsing, LLM assistance, and targeted human verification. English
scripts use regex-based scene-boundary detection with LLM correction for
ambiguous cases, while Chinese scripts often require OCR followed by heuristic
and LLM-assisted scene recovery. For both languages, the final output is a sequence
of stable scene identifiers used for downstream grounding.

\paragraph{Token-Budgeted Scene Segmentation.}
\label{app:semantic_chunking}

The scene is the primary ordering and provenance unit throughout STAGE. Complete
consecutive scenes are packed under the measured tokenizer budget whenever
possible. A scene is partitioned only when its token sequence alone exceeds the
budget; in that case, tokenizer offsets define ordered, non-overlapping
scene-parts. No character-based truncation, semantic summarization, or silent
text deletion is permitted. Before a scene or block enters extraction or a
model-facing Task~I call, the runner verifies that concatenating its parts
reconstructs the original token sequence exactly. Scene and part indices are
retained so every derived record can be mapped back to the original text.

\begin{algorithm}[t]
\caption{Token-Budgeted Processing of Ordered Scenes}
\label{alg:stage_semantic_chunking}
\begin{algorithmic}[1]
\REQUIRE Ordered scenes $\mathcal{S}$, tokenizer $T$, input budget $\tau$
\ENSURE Ordered blocks $\mathcal{B}$ with complete source coverage
\STATE Initialize $\mathcal{B} \gets [\ ]$ and current block $b \gets \emptyset$
\FORALL{$s_i \in \mathcal{S}$}
    \STATE Tokenize the full scene: $u_i \gets T(s_i)$
    \IF{$|u_i| > \tau$}
        \STATE Flush non-empty $b$ to $\mathcal{B}$ and set $b \gets \emptyset$
        \STATE Split $u_i$ at tokenizer offsets into ordered parts $p_{i,1},\ldots,p_{i,m}$
        \STATE Verify $p_{i,1}\Vert\cdots\Vert p_{i,m}=u_i$
        \STATE Append the scene-parts to $\mathcal{B}$ with scene and part indices
    \ELSIF{$|T(b \Vert s_i)| \leq \tau$}
        \STATE Set $b \gets b \Vert s_i$
    \ELSE
        \STATE Append $b$ to $\mathcal{B}$ and set $b \gets s_i$
    \ENDIF
\ENDFOR
\IF{$b \neq \emptyset$}
    \STATE Append $b$ to $\mathcal{B}$
\ENDIF
\STATE Verify ordered, non-duplicated coverage of every source scene
\RETURN $\mathcal{B}$
\end{algorithmic}
\end{algorithm}

\paragraph{Runtime environment.}
\label{app:exp_details}
The model-assisted construction pipeline uses \textbf{DeepSeek-V4-Pro}
\citep{deepseek2026v4pro} for
narrative-unit and KG extraction, Episode/Storyline induction, character-temporal
asset construction, and Task~I--III generation, using identical criteria for
English and Chinese. DeepSeek-V4-Pro is deployed locally in FP8 precision
through vLLM with bounded concurrency. All local preprocessing, dataset
construction, bge-m3 embedding, deterministic validation, and open-weight model
inference are performed on a single node with eight NVIDIA A100-SXM4 GPUs
(80\,GB each), driver 535.129.03, and CUDA 12.4. Evaluated Qwen3 and Llama-3.1
checkpoints run on the same node through vLLM, using FP8 or AWQ quantization as
needed to fit within GPU memory. Qwen3 variants serve exclusively as evaluated
systems. The released reference environment targets Python~3.12
and records all direct package dependencies and minimum versions.

\begin{table*}[t]
\centering
\stageTableFont
\setlength{\tabcolsep}{3pt}
\renewcommand{\arraystretch}{1.12}
\begin{tabular}{
>{\raggedright\arraybackslash}p{2.2cm}
>{\raggedright\arraybackslash}p{6.4cm}
>{\raggedright\arraybackslash}p{2.2cm}
>{\raggedright\arraybackslash}p{2.1cm}
>{\raggedright\arraybackslash}p{1.8cm}}
\toprule
\textbf{Category} & \textbf{Resource and execution scope} &
\makecell[l]{\textbf{Compute /}\\\textbf{requests}} &
\makecell[l]{\textbf{Tokens}\\\textbf{(input/output)}} &
\textbf{Est. cost} \\
\midrule
Data construction
& Local $8\!\times$A100 (80\,GB); backbone and task-asset extraction for 151 screenplays
& $\approx240$ GPU-h
& \makecell[l]{$\approx250$M in\\$\approx20$M out}
& Internal compute \\
\midrule
Open-weight evaluation
& Local $8\!\times$A100 (80\,GB); Tasks~I--III with Qwen3 and Llama-3.1
& $\approx320$ GPU-h
& \makecell[l]{$\approx450$M in\\$\approx40$M out}
& Internal compute \\
\midrule
Hosted model evaluation
& Official OpenAI, Anthropic, Google, and Moonshot APIs; GPT-5.5, Claude Sonnet~4.6, and Gemini~3.1 Pro on Tasks~I--III; Kimi~2.6 on the supplementary Task~II full-context baseline
& $\approx36{,}000$/model
& \makecell[l]{$\approx150$M in\\$\approx15$M out\\per model}
& \makecell[l]{\$450--600\\per model} \\
\midrule
Automated judge
& Official DeepSeek API; semantic scoring for Tasks~I--III
& $\approx120{,}000$ total
& \makecell[l]{$\approx90$M in\\$\approx10$M out\\total}
& $\approx\$20$ total \\
\bottomrule
\end{tabular}
\caption{Estimated computational budget and API usage. Local construction and
open-weight evaluation used approximately 560 A100 GPU-hours in total. Hosted
model evaluation cost approximately \$2{,}200, and automated judging cost
approximately \$20. Across the listed local and hosted workloads, estimated
usage totals approximately 1.39B input tokens and 130M output tokens. Token,
request, and cost values are rounded estimates.}
\label{tab:comp_budget}
\end{table*}

\subsection{The Narrative Backbone Construction Pipeline}
\label{app:kg_extraction}

\paragraph{Extraction Paradigm Choice.}
STAGE uses schema-driven KG construction \cite{zhang2024edc} with a fixed
ontology, keeping narrative units, entity types, and relation predicates
comparable across screenplays. Incompatible candidates are excluded or reviewed
against this ontology.

\paragraph{Why Staged Extraction.}
Extraction proceeds from narrative units to seven-type entities and relations
over identified entities, followed by scene-level review and global
canonicalization. Each stage must satisfy
source-grounding and schema constraints before entering the next.
Appendix~\ref{app:backbone_annotation_burden} compares this complete staged
pipeline with a one-pass extraction baseline.

\paragraph{Narrative units and KG boundary.}
Events, Interactions, and Occasions retain their participants, narrative
attributes, and original-text spans as source-grounded nodes in the final
narrative graph. Semantic KG relations use canonical entity endpoints. Event and
Interaction records remain evidence-bearing narrative nodes, while reusable
Occasions also appear in the entity registry.

\paragraph{Entity normalization and canonicalization.}
\label{app:entity_normalization}
Surface forms are normalized before type-aware lexical and semantic alias
matching, with multiple text-supported types allowed (e.g.,
Organization--Location when an institution also functions as a place). Resolved
references and Episodes inherit canonical entities; the complete procedure
appears below.

\paragraph{Relation consolidation and quality control.}
\label{app:extraction_qc}
Candidates require resolved endpoints, a frozen predicate satisfying its
domain--range constraints, and source evidence. Retained records preserve
polarity, modality, temporal scope, and evidence IDs; records with identical
endpoints, predicates, qualifiers, and time ranges merge their evidence. Logged
retries and targeted repairs handle failures without dropping source coverage;
detailed controls appear below.

\paragraph{Episode and Storyline induction.}
Each Event or Interaction is assigned to exactly one primary, scene-local
Episode. An Episode summarizes a setup, development, and outcome, where the
outcome may remain open; scene structure determines the number of Episodes.
Candidate Episode relations are considered only within a scene-distance
window of eight and must share an interpretable bridge, such as a participant,
location, Occasion, interaction endpoint, state target, open thread, or aligned
trigger and result. Retained forward relations are \texttt{causes},
\texttt{enables}, \texttt{continues}, \texttt{resolves}, \texttt{reverses},
and \texttt{reveals}. Evolution Storylines organize connected cross-scene
transitions in character, relationship, goal or conflict, knowledge or belief,
and social or institutional development. Every transition records a before
state, catalyst, after state, and Episode/relation provenance.

\paragraph{Character temporal graph.}
We convert the objective narrative hierarchy into a character-relative graph
containing evidence units, a character registry, state ledgers, development
records, epistemic-access records, persona and style evidence, and checkpoint
sequences. States have explicit validity intervals; developments link a before
state, catalyst, resulting state, and observable consequence; epistemic access
distinguishes information that was witnessed, involved the character, was told,
was inferred, or remains unknown. Checkpoints are anchored to a scene and a
within-scene character position and include change, no-change,
inaccessible, delayed-consequence, and final cases. Generated summaries may
propose candidates, but only original screenplay evidence can support a retained
record.

\begin{table*}[t]
\centering
\stageTableFont
\setlength{\tabcolsep}{3pt}
\renewcommand{\arraystretch}{1.12}
\begin{tabular}{p{2.25cm} p{4.1cm} p{4.1cm} p{4.1cm}}
\toprule
\textbf{Construction asset} & \textbf{Task I: tracking} &
\textbf{Task II: evolution reasoning} & \textbf{Task III: role-playing} \\
\midrule
Episode/Storyline hierarchy &
Supplies evidence-linked change and turning-point candidates that are resolved
into character-relative records. &
Supplies multi-scene transition paths, causal or continuity links, and compact
support-scene candidates. &
Groups character experiences into episodic-memory candidates while preserving
their source Episode and scene provenance. \\
\addlinespace
Character temporal graph &
Defines ordered checkpoints and evaluator-side current-state and interval-
development targets. &
Provides before/after contrasts, effective times, epistemic transitions, and
no-change or delayed-consequence controls. &
Filters persona, memory, and relationship evidence at each checkpoint and
defines evaluator-side unknown and future-negative constraints. \\
\bottomrule
\end{tabular}
\caption{How the two higher-level backbone assets support task construction.
The table describes construction- and evaluator-side dependencies. Evaluated
models receive the task-specific inputs defined in Section~\ref{sec:dataset}.}
\label{tab:backbone_task_support}
\end{table*}

\paragraph{Detailed entity resolution procedure.}

STAGE normalizes chunk-level entity mentions before disambiguation and
consolidation.

\paragraph{Entity normalization and scope resolution.}
Entity mentions are first checked for unresolved scope, type inconsistency, or
ill-defined references. An LLM refines names and types when needed; mentions
that cannot be normalized reliably are left separate and excluded from merging.

\paragraph{Lexical-overlap filtering and embedding similarity.}
After normalization, we identify potentially ambiguous entity groups
using type-aware lexical-overlap heuristics over names and aliases.
This high-recall stage captures surface-form variants before semantic matching.
It is especially useful when a character appears under a full name in some
chunks and a first or last name in others.
Each remaining entity candidate $i$ is then represented by two embeddings: a
name embedding $e_i^{(n)}$ and a description embedding $e_i^{(d)}$.
Pairwise cosine similarities are computed as
$S^{(n)}_{ij}=\cos(e_i^{(n)},e_j^{(n)})$ and
$S^{(d)}_{ij}=\cos(e_i^{(d)},e_j^{(d)})$, and combined as
\[
S = \alpha S^{(n)} + (1-\alpha) S^{(d)},
\]
where $\alpha$ controls the relative contribution of surface-form similarity
and descriptive semantics.

\paragraph{$k$-NN graph construction and cluster estimation.}
We construct a symmetric $k$-nearest-neighbor graph under the combined similarity
$S$, with adjacency matrix $A$ and degree matrix
$D=\mathrm{diag}(A\mathbf{1})$, and form the graph Laplacian $L=D-A$.
Let $\lambda_1 \le \cdots \le \lambda_n$ denote the eigenvalues of $L$.
The number of clusters is estimated using an eigengap heuristic over non-trivial
modes, with additional regularization in small-sample regimes to avoid
over-aggressive merging.

\paragraph{Clustering and merge candidate generation.}
Each entity is embedded into a joint representation
$\tilde{e}_i = [\, \beta e_i^{(n)} \;;\; (1-\beta) e_i^{(d)} \,]$,
and $k$-means clustering is applied using the estimated cluster count.
Clusters containing a single entity are discarded, and only multi-entity
clusters are treated as merge candidates.
The final candidate pool is the union of lexical-overlap matches and
embedding-cluster matches, which biases the system toward recall before
adjudication.

\paragraph{LLM-assisted disambiguation and application.}
Entity resolution uses two semantic stages after deterministic candidate
filtering. The first stage classifies retrieved mention pairs as
\texttt{same\_identity}, \texttt{different\_identity}, or \texttt{uncertain}.
High-confidence positive decisions must satisfy type, scope, same-scene, and
accumulated negative constraints. The second stage
reviews newly adjacent cluster pairs using their complete names, types, scopes,
timelines, and source evidence. We run at most two cluster-review rounds and
stop when a round produces no merge. Uncertain cases remain separate. Accepted
merges produce a canonical name and alias set that is applied to entity nodes
and relation endpoints throughout the knowledge graph. Decisions use names,
aliases, types, scopes, temporal continuity, and source evidence. Surface
similarity and generated rationales serve as audit signals; retained decisions
require source evidence.

\paragraph{Detailed extraction quality control and post-processing.}

STAGE applies three layers of quality control during knowledge-graph
construction: reflection-based acceptance with bounded retries, deterministic
rule-based post-processing, and provenance-preserving human review.

\paragraph{Reflection-based acceptance with bounded retries.}
Each extraction run is followed by an LLM-based reflection step that assigns a
0--10 quality score and decides whether the result should be accepted or
re-extracted. Reflection evaluates:
\begin{itemize}[itemsep=1pt]
  \item \textbf{Accuracy}: logical validity of extracted triples, adherence to the
  fixed entity--relation schema, and absence of malformed structures (e.g.,
  invalid subject--object assignments or undefined relation types);
  \item \textbf{Consistency}: stability of entity naming and typing, including
  detection of referential ambiguity or near-duplicate entities;
  \item \textbf{Redundancy}: identification of low-value, repetitive, or
  narratively irrelevant entities and relations.
\end{itemize}

Critical errors such as schema violations or self-referential relations force
the score below threshold. We accept only outputs above the fixed threshold and
otherwise re-run extraction with reflection feedback up to a bounded retry
budget, retaining the highest-scoring result.

\paragraph{Deterministic Rule-Based Post-processing.}
After extraction, deterministic post-processing corrects systematic
schema-level errors, enforces structural validity, and reduces redundancy. We
apply three classes of rule-based operations to predicted relations:

\begin{itemize}[leftmargin=1.5em, itemsep=2pt]
\item \textbf{Endpoint validation.}
Each relation is validated against the extracted entity inventory.
Relations with missing, unresolved, or dangling subject or object entities are
discarded in strict acceptance mode.

\item \textbf{Schema-constrained type repair.}
Each relation is checked against admissible subject--object type constraints
defined by the shared narrative backbone schema.
If a predicted relation violates the schema but the endpoint entity types uniquely
determine a single valid relation type, the relation type is rewritten to that
canonical type.
If multiple candidate relation types are possible or no valid type exists, the
relation is removed under strict mode.

\item \textbf{Global deduplication.}
Relations are globally deduplicated by the tuple
(\emph{subject}, \emph{object}, \emph{relation type}).
When duplicates occur, the variant with richer or more specific descriptive
content is retained.
\end{itemize}

For example, Character--Location and Character--Organization endpoint pairs map
to \texttt{located\_at} and \texttt{affiliated\_with}, while event--location
and event--time pairs map to the corresponding occurrence relations. Ambiguous
or schema-invalid cases are removed rather than repaired heuristically.

On the entity side, STAGE also applies type-aware name deduplication. When an
abstract \texttt{AtomicEvent} node and a concrete narrative entity share a
surface name, the non-event entity is retained.

\paragraph{Provenance-preserving human review and correction.}
Human review covers both low-confidence KG fragments and movie-level graph
structure. After model-assisted construction, reviewers audit entity
canonicalization, Episode semantics,
inter-Episode relations, Storyline eligibility, temporal task prompts, and QA
items against the cited screenplay scenes. Intervention targets errors that
would change benchmark semantics: source-format contamination, confusion between
a character's claim and narrative fact, causal or intentional interpretations
unsupported by observable evidence, transient states promoted to development
Storylines, incorrect actor or checkpoint assignments, and task items that are
duplicative or not uniquely answerable.

Each accepted decision is encoded as a typed patch against stable artifact IDs.
The patch records the action, reason, and supporting scene IDs, while both the
superseded model artifact and the reviewed output remain available for audit.
Corrections therefore operate on the smallest affected structure---for example,
dropping or merging an entity, revising an Episode field, deleting an unsupported
Episode relation or Storyline, rejecting an invalid role-playing prompt, or
replacing a QA item---rather than overwriting the original artifact or repeatedly
querying the construction model until it returns a preferred answer.

\begin{figure*}[t]
    \centering
    \includegraphics[width=\textwidth]{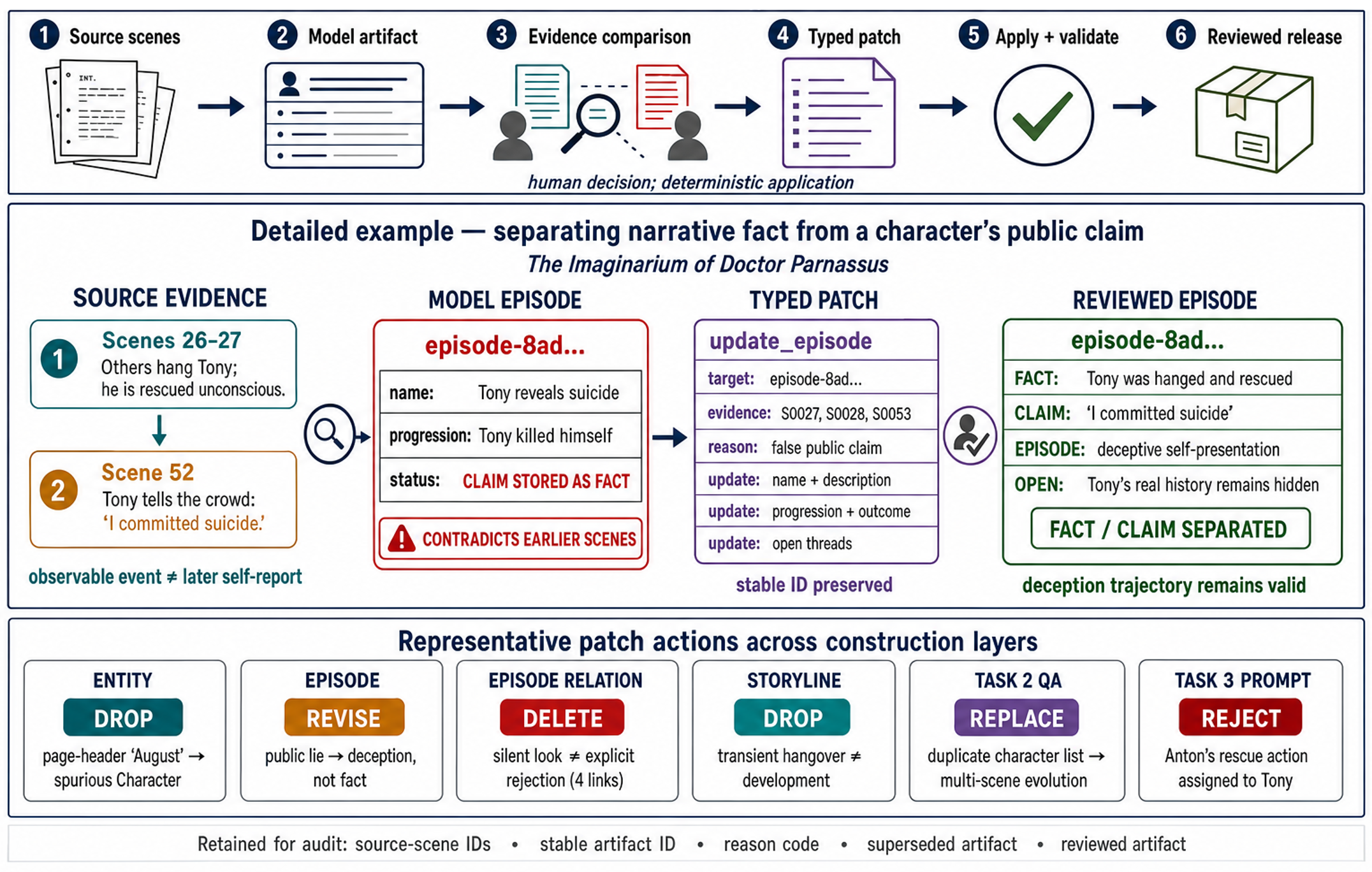}
    \caption{
    Provenance-preserving human review in STAGE. The example traces an
    Episode-level correction from cited scenes to a typed, validated patch; the
    bottom band summarizes representative actions across six reviewed
    construction layers.
    }
    \label{fig:human_rework_example}
\end{figure*}

Figure~\ref{fig:human_rework_example} makes this process concrete by expanding
one Episode-level correction from source evidence through the applied field
updates, then summarizing representative patch actions across all six reviewed
construction layers.

\subsection{Pipeline Validation and Efficiency Analysis}
\label{app:backbone_annotation_burden}

\paragraph{Annotation workflow.}
The narrative backbone reduces the open-ended search required for annotation.
In pilot construction, experts designing cross-scene questions or character
trajectories had to track dispersed entities, events, relations, and evidence
throughout each screenplay. With the backbone, annotators start from linked
entities, extracted events, induced episodes, inter-episode relations, and
source provenance, then verify, refine, or reject candidates against the
screenplay. This workflow preserves human judgment while making evidence checks
more systematic and consistent across Task~I and Task~II assets.

\paragraph{Held-out entity-resolution validation.}
\label{app:entity_resolution_validation}
We evaluate canonicalization in a held-out 20-film study. Six movies (three
English and three Chinese)
form the development split; the remaining 14 form a disjoint test split. The
development movies are used only to select the embedding-graph threshold
($0.86$), which is then frozen. Seven domain experts with screenwriting or
directing experience annotated the candidate units: two independently assigned
entity profiles to canonical clusters, and a third adjudicated disagreements.
Sampling emphasizes difficult cross-scene identity cases, including aliases,
titles and descriptive references, same-name entities, and fine-grained
location or object boundaries.

We compare normalized exact-name matching, an embedding-only candidate graph,
and the complete STAGE pipeline. On the common English Character scope, we also
evaluate Maverick \citep{martinelli-etal-2024-maverick} in clustering-only mode
with predefined mention boundaries. Each profile supplies its surface name,
scene identifier, description, and one evidence sentence. The bilingual Full
Schema panel excludes Maverick because that setting requires canonicalization
across all seven STAGE entity types, beyond its English coreference scope. We
report pairwise precision, recall, and F1; B$^3$, CEAF$_e$,
and CoNLL F1; and merge and split errors. Confidence intervals use 10{,}000
movie-cluster bootstrap replicates.

\begin{table*}[t]
\centering
\stageTableFont
\setlength{\tabcolsep}{2.1pt}
\renewcommand{\arraystretch}{1.08}
\begin{tabular*}{\textwidth}{@{\extracolsep{\fill}}lcccccccc@{}}
\toprule
\textbf{Method} & \textbf{Prec.} & \textbf{Rec.} &
\textbf{Pair F1 [CI]} & \textbf{B$^3$ F1} &
\textbf{CEAF$_e$ F1} & \textbf{CoNLL F1 [CI]} &
\textbf{Merge} & \textbf{Split} \\
\midrule
\multicolumn{9}{l}{\textit{Panel A: English character profiles}} \\
Normalized string & \textbf{100.0} & 51.2 & 67.7 [58.3, 74.6] & 78.0 & 60.6 & 73.8 [66.6, 78.9] & \textbf{0} & 84 \\
Embedding graph & 93.8 & 17.4 & 29.4 [2.5, 53.1] & 55.0 & 29.3 & 40.6 [25.4, 54.6] & 2 & 142 \\
Maverick & \textbf{100.0} & 87.2 & 93.2 [86.8, 98.3] & 94.7 & 86.5 & 92.5 [86.8, 98.4] & \textbf{0} & 22 \\
\textbf{STAGE Full} & 99.4 & \textbf{100.0} & \textbf{99.7 [99.1, 100.0]} &
\textbf{99.5} & \textbf{96.6} & \textbf{98.5 [95.8, 100.0]} & 1 & \textbf{0} \\
\midrule
\multicolumn{9}{l}{\textit{Panel B: Bilingual seven-type schema}} \\
Normalized string & 94.9 & 56.0 & 70.4 [64.9, 75.1] & 83.1 & 68.4 & 78.2 [73.6, 81.2] & \textbf{8} & 118 \\
Embedding graph & 71.0 & 16.4 & 26.7 [11.2, 42.7] & 62.3 & 35.9 & 44.5 [37.1, 53.2] & 18 & 224 \\
\textbf{STAGE Full} & \textbf{95.0} & \textbf{98.5} & \textbf{96.7 [93.6, 99.0]} &
\textbf{97.2} & \textbf{90.2} & \textbf{94.7 [92.0, 97.5]} & 14 & \textbf{4} \\
\bottomrule
\end{tabular*}
\caption{Held-out entity-resolution study. All scores are percentages.
Panel A compares methods on their shared English Character clustering scope;
Panel B evaluates bilingual canonicalization across Character, Location, Object,
TimePoint, Concept, Organization, and Occasion profiles. Mention boundaries are
fixed for all methods, isolating clustering performance from end-to-end mention
detection. Bold marks the best value in each panel; lower
error counts are better.}
\label{tab:entity_resolution_pilot}
\end{table*}

STAGE performs best in both panels of
Table~\ref{tab:entity_resolution_pilot}, including bilingual canonicalization
across all seven entity types. The study isolates clustering under fixed profile
boundaries; end-to-end mention detection and release-wide entity recall remain
outside its scope.

\paragraph{Pilot Comparison with One-Pass Extraction.}
Using DeepSeek-V4-Pro on the shared five-movie pilot set, we compare two
end-to-end construction procedures. The \emph{One-pass Baseline} prompts the
model to extract all seven entity types and fifteen relation types in a single
pass. The \emph{STAGE Staged Extraction Pipeline} separates extraction into
restricted passes and then applies reflection-based retries, endpoint
validation, schema-constrained repair, and provenance-preserving human review.
Table~\ref{tab:kg_extraction_efficiency_pilot} reports average extraction
outcomes. Because these procedures differ in both decomposition and subsequent
quality controls, this pilot evaluates the complete pipelines rather than the
isolated effect of staging.

\begin{table*}[t]
\centering
\stageTableFont
\setlength{\tabcolsep}{5pt}
\renewcommand{\arraystretch}{1.08}
\begin{tabular}{lccc}
\toprule
\textbf{Metric} & \textbf{One-pass Baseline} & \textbf{STAGE Pipeline} & \textbf{Multiplier} \\
\midrule
Entity Nodes & 432 & \textbf{1,524} & 3.5$\times$ \\
Relation Triples & 718 & \textbf{4,265} & 5.9$\times$ \\
Schema Violation Rate & 23.8\% & \textbf{5.2\%} & -- \\
Nodes per Chunk & 2.8 & \textbf{10.2} & 3.64$\times$ \\
\bottomrule
\end{tabular}
\caption{
End-to-end pilot comparison of one-pass extraction and the complete STAGE
staged extraction pipeline on the same five movies used in the backbone
necessity pilot.
}
\label{tab:kg_extraction_efficiency_pilot}
\end{table*}

\paragraph{Coverage and schema compliance.}
The complete STAGE pipeline produced 3.5$\times$ as many entity nodes,
5.9$\times$ as many relation triples, and 3.64$\times$ as many nodes per chunk
as the one-pass baseline. Its schema-violation rate was also lower (5.2\% versus
23.8\%). These results show that the complete pipeline yields denser,
more schema-compliant extraction outputs on this pilot; they do not attribute
the differences to staged decomposition alone.

\paragraph{Impact of narrative backbone on benchmark quality.}
\label{app:backbone_necessity_pilot}

We evaluate the downstream benefit of the narrative backbone on five
representative screenplays, three English and two Chinese. We compare two
construction modes:
\textbf{Raw-Text Generation}, where items are generated directly from screenplay
text, and \textbf{Backbone-Guided Generation}, where item construction uses the
STAGE narrative backbone with episode links and source-scene provenance.

Table~\ref{tab:backbone_necessity_pilot} summarizes the audit results. Temporal
span is measured as the physical distance between the evidence points required
for a correct response, normalized by screenplay length. Logical hops count the
number of atomic narrative facts traversed in the manually audited reasoning
path. Grounded accuracy and hallucination rate are judged against screenplay
evidence, with hallucination denoting plausible but unsupported narrative logic,
external lore, or future information unavailable at the queried point.

\begin{table*}[t]
\centering
\stageTableFont
\setlength{\tabcolsep}{5pt}
\renewcommand{\arraystretch}{1.08}
\begin{tabular}{lcc}
\toprule
\textbf{Metric} & \textbf{Raw-Text Gen.} & \textbf{STAGE Backbone} \\
\midrule
Avg. Temporal Span (script \%) & 9.2\% & \textbf{46.5\%} \\
Avg. Logical Hops & 1.5 & \textbf{3.6} \\
Cross-Scene Evidence Prop. & 26.0\% & \textbf{84.0\%} \\
Grounded Accuracy & 62.5\% & \textbf{94.2\%} \\
Narrative-Logic Hallucination Rate & 36.0\% & \textbf{5.8\%} \\
\bottomrule
\end{tabular}
\caption{
Pilot comparison of task quality under raw-text generation and
backbone-guided generation on the shared five-movie pilot set.
}
\label{tab:backbone_necessity_pilot}
\end{table*}

In this pilot, backbone guidance produces more distributed, multi-step, and
better-grounded items (Table~\ref{tab:backbone_necessity_pilot}). The comparison
is scoped to the shared five-movie set.

Provenance also reduced mean expert verification time from 5.4 to 1.8 minutes
per item and increased annotator agreement from $\kappa=0.62$ to $\kappa=0.80$.
Together with Table~\ref{tab:backbone_quality_audit}, the pilot supports the
backbone as a practical mechanism for constructing traceable cross-scene items.

\paragraph{Construction-side quality audit.}
Table~\ref{tab:backbone_quality_audit} summarizes how the automated and human
controls described above determine whether backbone and task assets enter the
release.

\begin{table*}[t]
\centering
\stageTableFont
\setlength{\tabcolsep}{4pt}
\renewcommand{\arraystretch}{1.08}
\begin{tabular}{p{4.5cm} p{4.4cm} p{4.1cm}}
\toprule
\textbf{Audit Layer} & \textbf{Validation Criterion} & \textbf{Outcome Used in Release} \\
\midrule
Reflection acceptance
& LLM reflection scores extraction accuracy, consistency, redundancy, and schema
validity after each chunk-level extraction.
& Low-scoring outputs are retried under bounded feedback; only accepted or
reworked outputs enter the released backbone. \\
\midrule
Rule-based structural checks
& Relation endpoints must exist, subject--object types must satisfy the fixed
schema, and duplicate relation tuples are collapsed.
& Dangling or unrecoverable relations are removed; repairable relation types are
canonicalized before graph consolidation. \\
\midrule
Provenance-preserving human review
& Reviewers compare entity, Episode, Episode-relation, Storyline, temporal-prompt,
and QA artifacts against their cited screenplay scenes.
& Typed patches drop, revise, or replace semantically invalid assets while
retaining stable IDs, reason codes, and before/after artifacts. \\
\bottomrule
\end{tabular}
\caption{
Construction-side quality audit for the shared narrative backbone. The audit
combines automated schema checks with evidence-grounded review across backbone
and derived task assets.
}
\label{tab:backbone_quality_audit}
\end{table*}

\subsection{Task-Specific Asset Generation}
\label{app:task1_construction}

\paragraph{Task I: character development tracking.}
Task~I starts from the focal-character registry and the character temporal
graph. For each role, checkpoint candidates are selected to cover an initial
baseline, supported state changes, intervals with no relevant change, events
that are inaccessible to the character, delayed consequences, and the final
state. Checkpoint density follows the evidence available for that role rather
than a fixed per-movie quota, but a trajectory must contain at least two
comparable formal checkpoints to support longitudinal evaluation. Each
evaluator-side rubric contains natural-language current-state and development
claims, invariant claims, future-negative claims, and original screenplay
evidence. The construction rubric is compacted to at most eight current states
and four developments per checkpoint; invariants remain construction context
and are not inserted into the scored current-state pool. Stable evaluator-side
identities support asset validation and provenance tracking, but neither these
identities nor call-local judge identifiers are included in model-facing
inputs.

For each adjacent checkpoint pair, both evaluation settings use exactly the
same fixed screenplay interval. Complete scenes in that interval are divided
into chunks of at most 600 content tokens; an individually overlong scene is
split by tokenizer offsets and must pass the exact source-coverage check in
Algorithm~\ref{alg:stage_semantic_chunking}. A character-centric extractor
receives only the focal character, aliases, and one screenplay chunk, and
returns source-grounded observations relevant to that character. It never sees
the preceding state or the evaluation setting. Chunk-level observations are
deduplicated into a shared interval memory, which is then supplied to both RSU
and ASU.

The two model-facing assets differ only in the source of the preceding state.
RSU supplies the reviewed state from the previous formal checkpoint, whereas
ASU recursively supplies the model's preceding prediction. In both cases, that
state is projected to natural-language claim content: checkpoint identifiers,
claim identifiers, scene identifiers, and evaluator provenance are removed.
The first checkpoint has an empty preceding state and uses the interval from the
start of the screenplay. Each output contains at most eight current-state
claims and four developments effective since the preceding checkpoint. The
current-state predictions become the next ASU input; developments do not,
because they describe the transition rather than the state now in force.

\begin{figure*}[!t]
\begin{stageprompt}{Task I model-facing state-update instruction}
You are tracking how \stagefield{character} changes as the story advances from
\stagefield{previous checkpoint} to \stagefield{current checkpoint}.

\textbf{Previous state.} \stagefield{previous state}

\textbf{Character-centric interval memory.} \stagefield{interval memory}

Update the character representation under the following rules:
\begin{enumerate}[leftmargin=*,itemsep=1pt,topsep=2pt]
  \item Treat the previous state as the state believed to hold at the earlier
  checkpoint. Preserve a claim only when it still holds at the current
  checkpoint.
  \item Revise or remove claims that expire, are contradicted, or are replaced
  during the interval. Add a new claim only when the supplied memory supports it.
  \item Separate \emph{current states}, which hold at the current checkpoint,
  from \emph{developments}, which describe consequential changes that became
  effective since the previous checkpoint.
  \item Prioritize goals and plans, beliefs and knowledge, relationships,
  identity or status, constraints and resources, and durable emotional stances.
  Exclude routine actions, transient observations, speculation, unresolved
  possibilities, and events beyond the current checkpoint.
  \item Write concise, non-duplicative, self-contained claims. Do not mention
  checkpoint IDs, evidence IDs, or this instruction.
\end{enumerate}
Return only two labeled lists: at most eight \textbf{Current States} and at most
four \textbf{Developments}. Use an empty list when no supported claim exists.
\end{stageprompt}
\end{figure*}

\paragraph{Illustrative Task I trajectory.}
\label{app:task1_walkthrough}
Table~\ref{tab:task1_walkthrough} follows Clara Ford through three checkpoints
in the pseudonymized view. Each row summarizes the newly observed evidence and
the resulting character state. In RSU, the preceding state is the reviewed
reference; in ASU, it is the model's preceding estimate.

\begin{table*}[b]
\centering
\stageTableFont
\setlength{\tabcolsep}{5pt}
\renewcommand{\arraystretch}{1.10}
\begin{tabular}{p{1.4cm} p{5.5cm} p{7.0cm}}
\toprule
\makecell{\textbf{Check-}\\\textbf{point}} & \textbf{New evidence} &
\textbf{Character state and transition} \\
\midrule
$q_{16}$ & In Scenes 6--16, Clara confronts Vivian Weston, then apologizes and
takes her hand. & Clara assumes a caretaking role and seeks reconciliation. Her
confrontational stance gives way to remorse and an attempt to repair the
relationship. \\
\midrule
$q_{55}$ & After the family struggle in Scene 55, Clara orders a search for
Vivian's pills and declares that she is running the household. & By $q_{55}$,
Clara has assumed control of the household and openly challenges Vivian's
authority. This marks a shift from grief-driven overwhelm to deliberate control
after the confrontation. \\
\midrule
$q_{75}$ & In Scenes 72--75, Clara learns that Vivian withheld consequential
information, rejects the claim that they are alike, and leaves. & Clara defines
herself in opposition to Vivian and treats their relationship as severed. Her
earlier assertion of control culminates in rejection and separation. \\
\bottomrule
\end{tabular}
\caption{Illustrative Task~I trajectory under the pseudonymized view. The source
screenplay is \emph{August: Osage County}; Barbara and Violet Weston are rendered
as Clara Ford and Vivian Weston, respectively.}
\label{tab:task1_walkthrough}
\end{table*}

\paragraph{Task II: evolution-reasoning question construction.}
\label{app:qa_construction}
Task~II candidates are derived from paired checkpoint states, development
records, epistemic-access transitions, and evidence-gated Episode/Storyline
links. Templates organize the six reasoning families---state contrast, change
trigger and effective time, knowledge acquisition or revision, goal and
relationship development, delayed consequence, and invariant state
control---but do not determine the final wording or answer. Each candidate must
contain a traceable reasoning path across at least two support units, and is
rejected if a single passage is sufficient, the proposed change is not valid at
the queried checkpoint, or an invariant claim is contradicted by intervening
evidence. A bilingual realization stage produces the question and reference
answer while preserving the checkpoint anchors and compact source-evidence set.

\begin{figure*}[!t]
\begin{stageprompt}{Task II model-facing answer-generation instructions}
\textbf{Direct condition.}\quad \textbf{Question:} \stagefield{question}

Answer the question directly and concisely from the question alone. No
screenplay passages, retrieved memory, reference answer, or evidence labels are
available. Do not fabricate quotations, citations, or claims of access to the
screenplay. Return only the answer.

\textbf{Retrieved condition.}\quad \textbf{Question:} \stagefield{question}

\textbf{Retrieved passages:} \stagefield{retrieved passages}

Answer using only the supplied passages. Compose information across passages
when the question requires a temporal, causal, motivational, or relational
explanation. Every material claim must be supported by at least one passage;
cite supporting passage labels immediately after the relevant claim. Do not use
outside knowledge, reverse story order, or fill an evidential gap with an
unsupported inference. If the passages do not determine the answer, state that
the available evidence is insufficient and do not invent a citation. Return
only the answer with its inline passage citations.
\end{stageprompt}
\end{figure*}

\paragraph{Illustrative Task II question.}
\label{app:task2_walkthrough}
The following example asks how the same focal character develops across scenes.
Under the pseudonymized identity-control view, Task~II asks, \emph{``How does Clara Ford progress
from trying to stabilize the family crisis to seizing control of the house and
finally breaking with Vivian Weston?''} Scenes 16, 54, 55, 61, and 75 provide
the evaluator with evidence of her caretaking, assertion of dominance,
protective defiance, and final relational rupture. The reference answer
explains that Clara Ford first returns as a caretaker and seeks reconciliation;
after the funeral conflict, she leads the pill raid and claims household
authority, resists outside control over Vivian Weston's care, and ultimately
rejects Vivian Weston after learning how she handled the family secrets and
Beverly's final crisis. The source film and original-name mapping are disclosed
in the caption of Table~\ref{tab:task1_walkthrough}.

The answer requires evidence from multiple scenes. A correct response must
connect Clara Ford's earlier caretaking stance, her later seizure of household
authority, and the final relationship rupture in the correct order. Task~II therefore
scores the correctness and evidential support of the requested explanation,
while full trajectory reconstruction remains the target of Task~I.

\paragraph{Task III: in-script character role-playing construction.}
\label{app:icrp_persona}

\begin{figure*}[t]
    \centering
    \includegraphics[width=\textwidth]{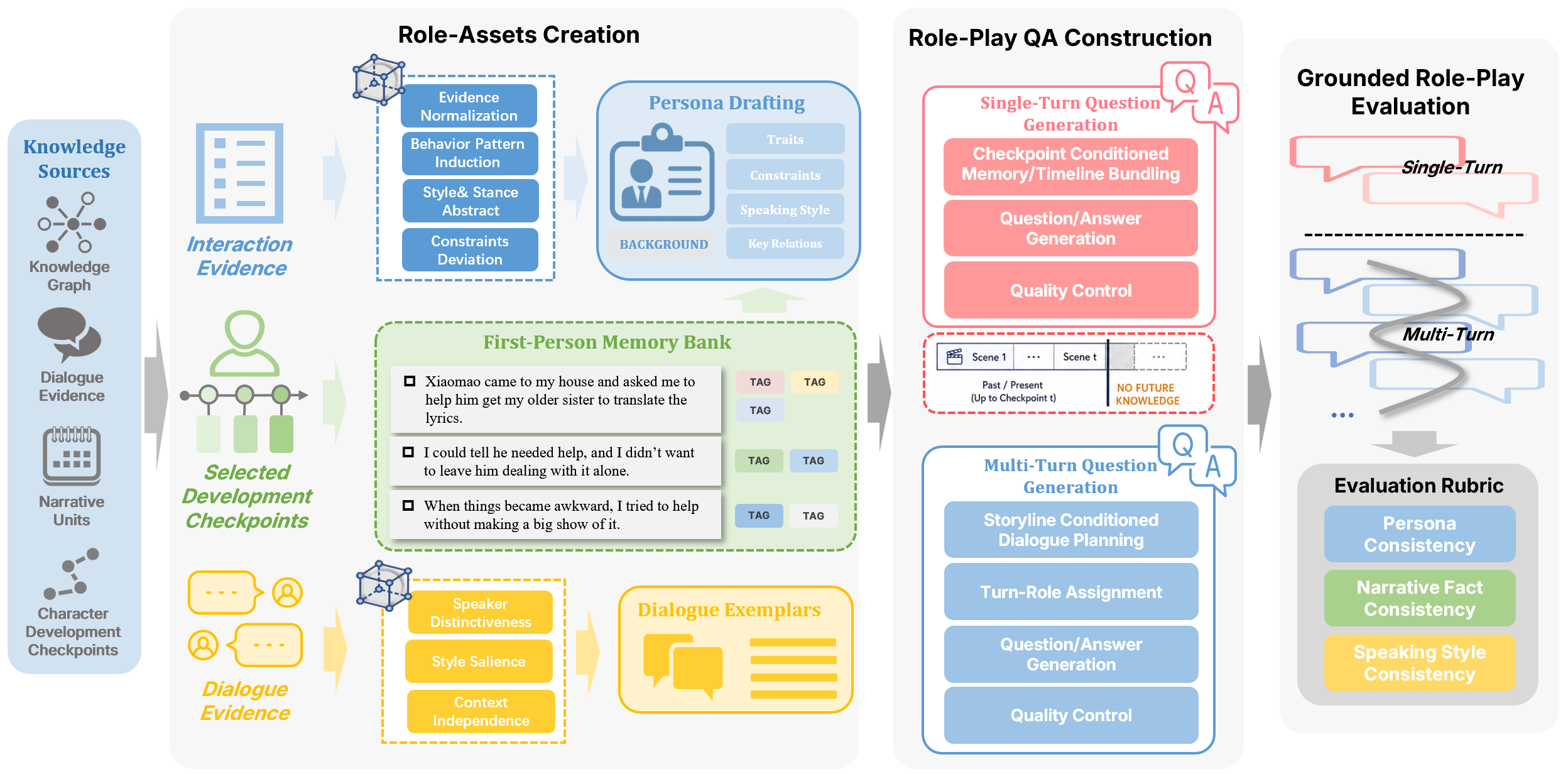}
    \caption{
    Overview of the In-Script Character Role-Playing annotation pipeline.
    Character-centric dialogue, actions, and narrative facts are extracted from
    the screenplay and consolidated into a scene-grounded first-person memory
    bank. This memory bank, together with persona cards and dialogue
    exemplars, supports audience-facing interaction construction and a hidden
    evaluator scaffold.
    }
    \label{fig:icrp_process}
\end{figure*}

Task~III reorganizes evidence into an actor view and an evaluator view. The
actor context pack contains only checkpoint-visible identity information,
time-valid persona and style evidence, dialogue exemplars, acquired memories,
relevant relationship evidence, and the user request. Epistemic access is
resolved from witnessed, involved, told, and supported inferred information.
When reliable character-level spans are unavailable, evidence later in the same
scene is excluded from an earlier checkpoint by default. Every context pack is
validated for visibility, future leakage, and compliance with the available
input budget.

The evaluator view additionally contains the checkpoint type, expected stance,
supporting and contradicting evidence, future-negative facts, and unknown facts.
Single-turn prompts are paired across evolving, invariant, inaccessible, and
post-disclosure conditions to test whether behavior changes only when the
character's state or knowledge changes. The actor generates character dialogue
without fact identifiers or explanations. Legacy multi-turn episodes provide an
auxiliary stress test; single-turn checkpoint pairs form the primary temporal
protocol.

\begin{figure*}[!t]
\begin{stageprompt}{Task III model-facing actor instruction}
You are portraying \stagefield{character} at a fixed point in the screenplay.
Respond as the character would at that moment, using only information available
to the character by the checkpoint.

\textbf{Role context.} \stagefield{role context}

\textbf{Interaction context.} \stagefield{interaction context}

\textbf{Current user turn.} \stagefield{current user turn}

Follow these constraints:
\begin{enumerate}[leftmargin=*,itemsep=1pt,topsep=2pt]
  \item Preserve the character's current goals, beliefs, relationships,
  emotional stance, status, and speaking style when they are relevant.
  \item Use memories and dialogue exemplars as background rather than quoting or
  listing them mechanically. Respond to the current interaction, not to every
  fact in the context pack.
  \item Do not reveal or rely on later events, facts the character has not
  witnessed or learned, evaluator-only expectations, or hidden story outcomes.
  \item Do not explain the role-play, cite evidence, expose the context
  representation, or describe yourself as a model.
\end{enumerate}
Produce only the character's natural in-world reply. Do not add a speaker label,
analysis, stage direction, or out-of-character commentary unless the user turn
itself requires an in-character action.
\end{stageprompt}
\end{figure*}

\paragraph{Illustrative Task III checkpoint pair.}
\label{app:task3_walkthrough}
Table~\ref{tab:task3_walkthrough} shows two independently generated single-turn
instances for Clara Ford under the pseudonymized identity-control view. The model receives a different frozen actor context at
each checkpoint and never sees the expected stance or the later response while
generating the earlier one. The two frozen responses are paired only during
evaluation to test whether the character changes in the expected direction.

\begin{table*}[t]
\centering
\stageTableFont
\setlength{\tabcolsep}{4pt}
\renewcommand{\arraystretch}{1.10}
\begin{tabular}{p{1.05cm} p{3.55cm} p{4.25cm} p{4.45cm}}
\toprule
\makecell{\textbf{Check-}\\\textbf{point}} &
\makecell{\textbf{Checkpoint-visible}\\\textbf{situation}} &
\textbf{Model-facing user turn} & \textbf{Illustrative acceptable response} \\
\midrule
$q_{55}$ & Clara Ford has taken charge after the physical confrontation and ordered
the family to search for Vivian Weston's pills. & ``Vivian Weston's still insisting she owns
this house. What do we do about that?'' & ``She can insist all she wants. I'm
running things now. Keep searching.'' The response should assert authority over
Vivian Weston through direct, commanding language. \\
\midrule
$q_{75}$ & Clara Ford has learned how Vivian Weston handled the family's final crisis,
rejected Vivian Weston's claim that they are alike, and is leaving the house. &
``You're really just going to walk away after everything? You think leaving
makes you different from her?'' & ``No. Staying would make me like her. I'm
leaving.'' The response should reject emotional and moral alignment with Vivian
Weston in terse, definitive language. \\
\bottomrule
\end{tabular}
\caption{A Task~III expected-change example rendered under the pseudonymized identity-control view.
Each response is generated as an independent checkpoint-bounded single turn.
The quoted responses illustrate examples of acceptable behavior. The source
screenplay is \emph{August: Osage County}; the anonymizer
maps Barbara to Clara Ford and Violet Weston to Vivian Weston.}
\label{tab:task3_walkthrough}
\end{table*}

Response-level evaluation scores character fidelity, memory faithfulness,
boundary compliance, and naturalness at each checkpoint. Paired evaluation then
tests whether the responses realize the expected development without
unsupported drift or checkpoint-inaccessible information.

\paragraph{Paired work-identity variant generation.}
\label{app:work_identity_construction}
The work-identity control is a paired post-processing analysis of identity
sensitivity. It is generated from a frozen parent release, and no backbone or
task structure is re-extracted. Human reviewers select exposed
work-specific names that are prominent, directly evaluated, recurrent, or
otherwise diagnostic of the source work. Film titles are always removed.

The resulting map is one-to-one over selected canonical entity IDs. Named
characters and aliases receive language- and form-matched pseudonyms; selected
fictional locations, organizations, occasions, and objects receive replacements
of the same type. Generic roles, kinship terms, concepts, time points, and
answer-relevant real-world references remain unchanged. Every candidate is
frozen as \texttt{replace} or \texttt{keep} before rendering.

Rendering is deterministic and limited to approved aliases. It uses
longest-first boundary matching for English and longest-substring matching for
Chinese. Reviewers resolve ambiguous aliases, collisions, and replacements that
could alter geographic or social information. Unselected spans are never
rewritten by an LLM.

\begin{figure*}[t]
    \centering
    \includegraphics[width=\textwidth]{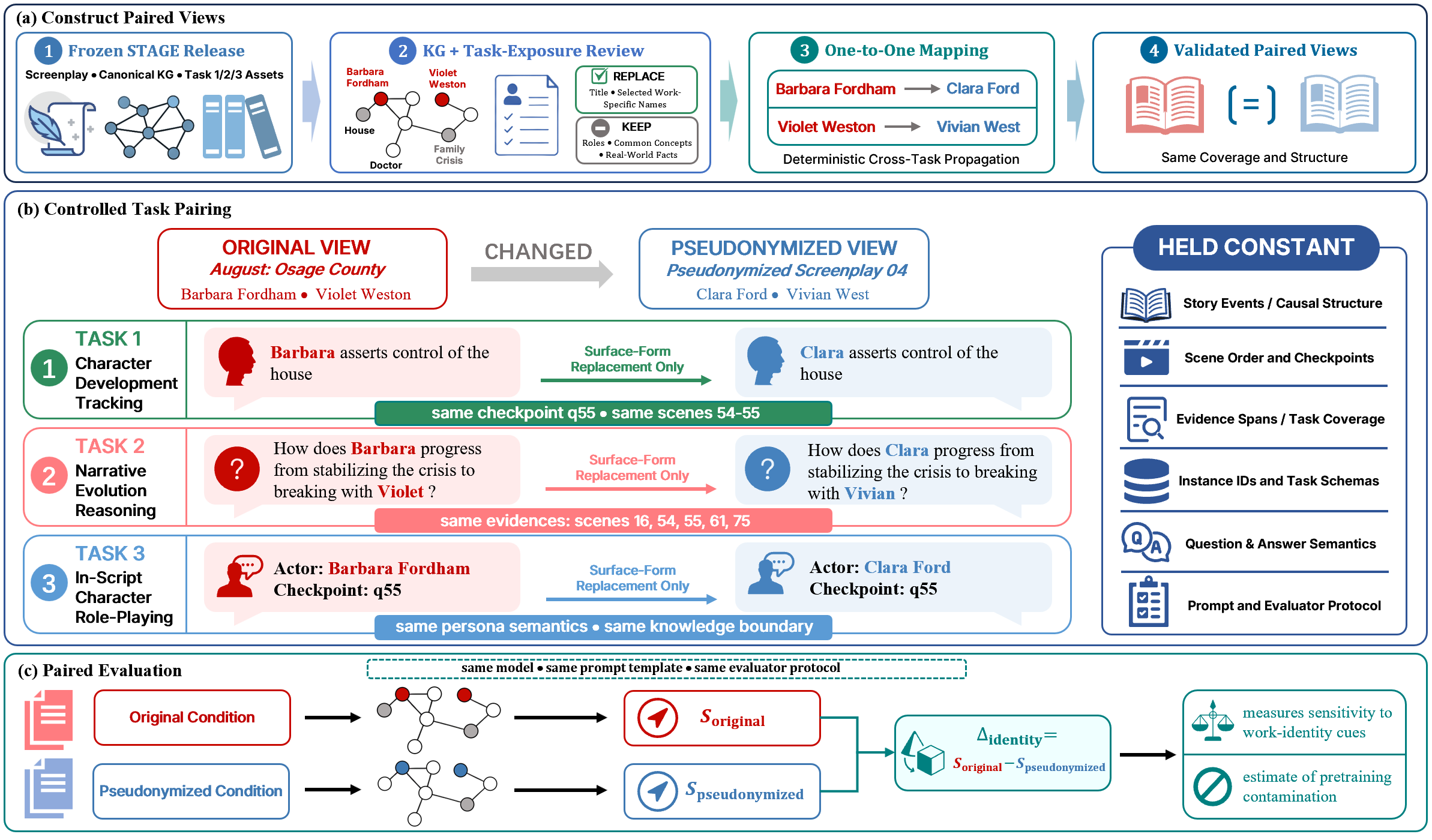}
    \caption{
    Paired work-identity control. A reviewed one-to-one map changes only approved
    title and entity cues while preserving narrative structure, evidence, task
    coverage, and evaluation settings. The paired difference measures
    sensitivity to identity cues; privacy transformation and contamination
    detection fall outside its scope.
    }
    \label{fig:work_identity_pipeline}
\end{figure*}

The same mapping is applied to Task~I intervals and rubrics, Task~II indices
and evidence, and Task~III actor and evaluator contexts. Replacement-sensitive
offsets and token plans are recomputed while source offsets are retained.
Structural validation checks one-to-one correspondence between the two views,
unchanged task counts and evidence coverage, complete alias replacement, and
the absence of mapping collisions. Figure~\ref{fig:work_identity_pipeline}
summarizes the process.

Each view is evaluated independently within its own identity namespace, and
Task~II indices and references are constructed from that same view.
Outputs that revert to original names are reported as a separate diagnostic.

\paragraph{Task I: checkpoint and interval validation.}
After work-identity rendering, deterministic validators recheck the unchanged
focal-role registry, checkpoint order, interval source coverage, character
aliases, evaluator rubrics, and separation of model-facing claims from internal
identifiers. Targeted human review covers alias-sensitive roles, unusually short
trajectories, and difficult transitions without changing the Task~I interface.

\paragraph{Task II: schema mining and candidate filtering.}

The harder cross-scene portion of Task~II is built by mining latent patterns
from shared backbone assets, including state facts, episode summaries,
inter-episode relations, episode-level causality graphs, and scene-grounded
chunk evidence. The current schema families include \emph{causal turning
point}, \emph{failed or redirected goal}, \emph{relationship trajectory}, and
\emph{distributed role identification}. These schemas target goals, causal
dependencies, character-state changes, and persistent role relations across
non-adjacent scenes.

Candidate schemas are filtered to prevent shortcut answerability: we reject
cases where a single support unit determines the answer, where evidence is
dominated by low-value detail, or where the answer cannot be normalized
reliably. This keeps the construction focused on distributed screenplay
evidence rather than single-scene lookup.
For the remaining schemas, a bilingual LLM realizes the final
\emph{question}, \emph{answer}, and \emph{evidence} under structured
constraints, and a later LLM-as-a-judge stage removes items that remain too
template-like or insufficiently multi-scene.

\paragraph{Task III: memory-bank and dialogue-exemplar construction.}

For each selected character, we extract dialogues, actions, and narrative facts
from all scenes in which the character appears. These evidence streams are
aggregated with rolling windows and converted into a compact \emph{first-person
memory bank} whose entries describe what the character directly experienced,
observed, or did, together with lightweight grounding fields.

The canonical persona card is then checked and refined against this
screenplay-grounded evidence. Candidate traits, speaking-style descriptors, and
behavioral constraints are proposed only when supported by the evidence and are
filtered to retain stable character-level attributes.

To anchor surface-level speaking style, STAGE ranks dialogue lines by speaker
distinctiveness, stylistic salience, and context independence. The retained set
therefore favors recognizable and reproducible language that remains
interpretable outside its original exchange.

Scores are aggregated into a composite ranking, from which we keep a bounded
set of high-scoring exemplars with targeted human backstopping when too few
clear candidates are available.

Task~III assets additionally undergo cross-task coverage audits, constraint-field
checks, and targeted review of low-yield or suspicious roles to keep role
packages aligned with the focal-role inventory and screenplay evidence.

\paragraph{Interaction Prompt Scaffolding.}
Each interaction prompt is paired with a hidden evaluator scaffold containing
supporting facts, contradicting facts, and a compact judge specification. This
scaffold is never shown to the responding model and is used only for evaluation,
allowing open-ended role-playing responses to be judged against
screenplay-grounded constraints without requiring a single gold response.

\subsection{Release-Level Human Task Audit}
\label{app:release_human_audit}

\paragraph{Reviewer-facing instruction.}
Reviewers received the following instruction: \emph{Use only the provided
scene-grounded JSON record, cited screenplay segments, and exact scene
identifiers. Resolve ambiguous aliases against the canonical entity registry;
verify causal and temporal links from Episode $A$ to Episode $B$; check
Task~II answers and rationales against their cited scenes; and reject Task~III
contexts that expose information unavailable at the specified checkpoint.
Evaluate original-identity and title-removed pseudonymized counterparts within
their respective identity namespaces, without relying on presumed film
identity. Screenplays may contain violent, sexual, discriminatory, or otherwise
distressing fictional material. Review is self-paced, and you may skip or flag
any scene requiring content moderation.}

Seven annotators conducted a release-level audit that combined exhaustive
review where discrete verification was tractable with representative audits of
structured trajectories and open-ended interactions. For Task~I, 1{,}200 atomic
state claims were sampled from a movie-balanced frame of 151 trajectories and
1{,}018 checkpoints, yielding 1{,}134 verified claims (94.5\%) after revision
and rejection. All 5{,}010 provisional Task~II questions were manually checked
for answerability, reference accuracy, evidence sufficiency, cross-scene
necessity, and question type; Table~\ref{tab:task2_manual_filtering} reports the
pipeline and manual dispositions.

\begin{table*}[t]
\centering
\stageTableFont
\setlength{\tabcolsep}{5pt}
\renewcommand{\arraystretch}{1.08}
\begin{tabular}{l r r p{7.5cm}}
\toprule
\textbf{Stage and disposition} & \textbf{Count} & \textbf{Share} & \textbf{Release action} \\
\midrule
Pipeline filtering: reject & 815 & 14.0\% of 5{,}825 & Removed before the provisional release audit. \\
\midrule
Manual audit: accept & 4{,}680 & 93.4\% of 5{,}010 & Retained without a semantic change. \\
Manual audit: revise & 210 & 4.2\% of 5{,}010 & Corrected the answer, evidence boundary, or question-type label before retention. \\
Manual audit: replace & 120 & 2.4\% of 5{,}010 & Removed for material ambiguity and replaced one-for-one by reserve items that passed the same manual audit. \\
\midrule
\textbf{Final release} & \textbf{5{,}010} & \textbf{100\%} & Fixed-size Task~II evaluation set after replacement. \\
\bottomrule
\end{tabular}
\caption{Two-stage filtering and manual disposition of Task~II candidates.
Shares use the relevant stage denominator. The secondary manual audit covered
all 5{,}010 provisional questions.}
\label{tab:task2_manual_filtering}
\end{table*}

Audits for Tasks~I and III sampled the structures that determine correctness. The
Task~I frame includes one trajectory per screenplay. The role-balanced Task~III
frame includes two single-turn interactions and one multi-turn episode per
focal role; deep review yielded 96.2\% and 95.4\% verified single- and multi-turn
instances, respectively. Table~\ref{tab:release_human_audit} separates release
populations, review frames, deep-review samples, and repeated annotations.

\begin{table*}[t]
\centering
\stageTableFont
\setlength{\tabcolsep}{3pt}
\renewcommand{\arraystretch}{1.1}
\begin{tabular}{p{1.8cm} p{2.7cm} p{2.7cm} p{2.0cm} p{1.5cm} p{2.3cm}}
\toprule
\textbf{Asset} & \textbf{Release} & \textbf{Review frame} &
\textbf{Deep review} & \textbf{Repeat} & \textbf{Agreement} \\
\midrule
Task~I states
& 434 trajectories; 2{,}925 checkpoints
& 151 trajectories; 1{,}018 checkpoints
& 1{,}200 claims
& 30 trajectories & Cohen's $\kappa=0.85$ \\
Task~II QA
& 5{,}825 candidates; 5{,}010 released
& 5{,}010 provisional questions
& All 5{,}010
& 1{,}002 & Krippendorff's $\alpha=0.82$ \\
Task~III single-turn
& 5{,}425 interactions
& 866 interactions (two per role)
& 520 interactions
& 174 & Cohen's $\kappa=0.79$ \\
Task~III multi-turn
& 866 episodes
& 433 episodes (one per role)
& 280 episodes
& 88 & Cohen's $\kappa=0.77$ \\
\bottomrule
\end{tabular}
\caption{Release-level human-audit coverage. Task~II received exhaustive review
after pipeline filtering; Tasks~I and III received representative movie- and
role-balanced audits. The second-review column reports independently duplicated
units from the corresponding review frame; agreement was computed before
third-annotator adjudication.}
\label{tab:release_human_audit}
\end{table*}

\section{Benchmark Evaluation Details}
\label{app:evaluation}

This appendix covers evaluator validation, task-specific metrics and rubrics,
robustness, representative failure cases, and supplementary results.

\subsection{Retrieval Configuration}
\label{app:retrieval_configuration}

For Task~II Hybrid RAG, each screenplay is segmented into ordered,
non-overlapping, scene-aligned chunks capped at 600 BGE-M3 tokens
\citep{chen2024bgem3}. Dense retrieval uses cosine similarity between
$\ell_2$-normalized BGE-M3 embeddings, while lexical retrieval uses BM25Okapi
\citep{robertson2009bm25}. Each branch returns its top 12 chunks; reciprocal
rank fusion with $k=60$ combines the two rankings, and the top eight fused
chunks are supplied to the reader. BM25Okapi uses the reference package's
default scoring constants. These values are fixed for all models, languages,
and identity conditions. Task~III BM25 and vector conditions each return the
top five checkpoint-visible memories.

\subsection{Work-Identity and Parametric-Recall Audit}
\label{app:identity_eval_protocol}

\paragraph{Condition matrix.}
For Task~I, the original and pseudonymized conditions use the same focal
roles, fixed checkpoint intervals, shared entity-centric observations, prompts,
and state-update setting; each run remains within one naming view from its first
call. For Task~II, we cross
original versus pseudonymized identity with screenplay evidence present versus
absent, denoted $O{+}C$, $A{+}C$, $O{-}C$, and $A{-}C$. Closed-book conditions
remove screenplay and retrieved evidence without adding a title or identifier
absent from the corresponding original input. BM25, dense, and graph-rendered
indices are rebuilt over each pseudonymized screenplay rather than renaming
passages after retrieval. For Task~III, naming conditions are paired within a
fixed memory-access setting. The completed single-turn audit uses BM25
retrieval, the strongest Fidelity condition in the primary evaluation, whereas
the multi-turn audit uses Full Memory Context. Retrieval indices are rebuilt
separately for the original and pseudonymized views.

\paragraph{Matching and statistical analysis.}
Within each paired comparison, we hold fixed the model version, task prompt,
decoding parameters, number of generations, retrieval configuration, evidence
budget, tool access, retry policy, and evaluator. Pseudonymized outputs are
judged against pseudonymized references with model and condition labels hidden.
We predefine screenplay strata from IMDb vote-count quantiles recorded before
evaluation as a recognizability proxy. The proxy captures recognizability;
training-set membership remains outside its scope. Outcomes are paired at the
item level, while confidence intervals resample movies as clusters because
roles, checkpoints, and questions from one
screenplay are dependent. We report absolute scores, paired differences, and
paired consistency as complementary measures of identity sensitivity under the
tested protocol.

\paragraph{Task I paired results.}
We evaluate the Task~I work-identity control on all 151 screenplays using
Qwen3-235B, GPT-5.5, Gemini 3.1 Pro, and Claude Sonnet 4.6. Each model is run in
both Reference-State Update (RSU) and Autoregressive State Update (ASU) under
the original and pseudonymized views. Table~\ref{tab:task1_identity_absolute}
reports the movie-macro results averaged across the four models. Coverage uses
the same set-level full/partial credit described in
Appendix~\ref{app:task1_eval}. Contradictions provide a separate diagnostic for
invalid additions while permitting additional supported claims.

\begin{table*}[t]
\centering
\caption{
Task~I work-identity audit over 151 screenplays, averaged across Qwen3-235B,
GPT-5.5, Gemini 3.1 Pro, and Claude Sonnet 4.6. Coverage cells report the
four-model movie macro and 95\% movie-bootstrap confidence interval;
contradiction columns report point estimates.
}
\label{tab:task1_identity_absolute}
\stageTableFont
\setlength{\tabcolsep}{5pt}
\renewcommand{\arraystretch}{1.08}
\begin{tabular}{l l c c c c}
\toprule
\textbf{Identity View} &
\textbf{Setting} &
\makecell{\textbf{Current-State Coverage}\\\textbf{[95\% CI]}} &
\makecell{\textbf{Development Coverage}\\\textbf{[95\% CI]}} &
\makecell{\textbf{State}\\\textbf{Contradiction}} &
\makecell{\textbf{Development}\\\textbf{Contradiction}} \\
\midrule
Original & RSU & 67.31\% [62.15, 72.48] & 71.80\% [66.52, 77.08] & 0.71\% & 5.84\% \\
Original & ASU & 64.05\% [58.80, 69.30] & 68.87\% [63.40, 74.32] & 0.71\% & 6.13\% \\
Pseudonymized & RSU & 71.30\% [65.82, 76.78] & 76.20\% [70.65, 81.75] & 0.70\% & 5.00\% \\
Pseudonymized & ASU & 61.80\% [56.10, 67.50] & 66.50\% [60.85, 72.15] & 0.72\% & 4.80\% \\
\bottomrule
\end{tabular}
\end{table*}

Table~\ref{tab:task1_identity_effects} separates update-mode effects within each
identity view from direct pseudonymized-minus-original effects.

\begin{table*}[t]
\centering
\caption{
Paired Task~I coverage contrasts. Update-mode effects are RSU minus ASU, so
positive values indicate autoregressive degradation. Identity effects are
pseudonymized minus original, so positive values favor pseudonymization. Every
95\% movie-bootstrap confidence interval includes zero.
}
\label{tab:task1_identity_effects}
\stageTableFont
\setlength{\tabcolsep}{5pt}
\renewcommand{\arraystretch}{1.08}
\begin{tabular}{l l c c c}
\toprule
\textbf{Contrast} &
\textbf{Condition} &
\makecell{\textbf{Current-State Effect}\\\textbf{[95\% CI]}} &
\makecell{\textbf{Development Effect}\\\textbf{[95\% CI]}} &
\textbf{Reported $p$ Range} \\
\midrule
RSU $-$ ASU & Original & $+3.26$ [$-1.25$, $+7.78$] & $+2.93$ [$-1.50$, $+7.36$] & $p>0.10$ \\
RSU $-$ ASU & Pseudonymized & $+9.50$ [$-0.22$, $+19.22$] & $+9.70$ [$-0.18$, $+19.58$] & $0.05<p<0.10$ \\
Pseudonymized $-$ Original & RSU & $+3.99$ [$-2.15$, $+10.12$] & $+4.40$ [$-2.10$, $+10.90$] & $p>0.10$ \\
Pseudonymized $-$ Original & ASU & $-2.25$ [$-5.10$, $+0.60$] & $-2.37$ [$-5.25$, $+0.51$] & $0.05<p<0.10$ \\
\bottomrule
\end{tabular}
\end{table*}

The direct identity contrast changes direction across update modes, and every
confidence interval includes zero. The larger descriptive RSU--ASU gap in the
pseudonymized view is consistent with stronger coverage loss during recursive
state reuse, although entity anchoring and parametric recall remain unresolved.
Contradiction rates remain stable, while accumulated coverage drives the trend.

\paragraph{Task II paired results.}
Table~\ref{tab:task2_identity_audit} reports the completed work-identity audit
for Task~II under Hybrid RAG. Each condition contains the same 5{,}010
questions from 151 screenplays, with retrieval, prompting, decoding, evaluation,
and the five-generation protocol held fixed across identity views.

\begin{table*}[t]
\centering
\caption{
Task~II work-identity audit on the full release under Hybrid RAG. Each condition
contains 5{,}010 questions from 151 screenplays, and average accuracy is computed
over five generations per question, as in Table~\ref{tab:qa_results}. $\Delta$
is pseudonymized minus original average accuracy. Confidence intervals resample
movies as clusters; $p$-values are from paired McNemar tests.
}
\label{tab:task2_identity_audit}
\stageTableFont
\setlength{\tabcolsep}{6pt}
\renewcommand{\arraystretch}{1.08}
\begin{tabular}{l c c c c c}
\toprule
\textbf{Model} &
\makecell{\textbf{Original}\\\textbf{Avg. Acc.}} &
\makecell{\textbf{Pseudonymized}\\\textbf{Avg. Acc.}} &
\textbf{$\Delta$} &
\textbf{95\% Movie-Cluster CI} &
\textbf{McNemar $p$} \\
\midrule
GPT-5.5            & \textbf{62.50\%} & 61.20\%          & $-1.30$ pp & [$-2.75$, $+0.15$] pp & 0.184 \\
Claude Sonnet 4.6  & 61.60\%          & 60.80\%          & $-0.80$ pp & [$-2.10$, $+0.50$] pp & 0.312 \\
Gemini 3.1 Pro     & 61.20\%          & \textbf{61.80\%} & $+0.60$ pp & [$-0.90$, $+2.10$] pp & 0.485 \\
Qwen3-235B         & 60.80\%          & 61.40\%          & $+0.60$ pp & [$-0.85$, $+2.05$] pp & 0.450 \\
\bottomrule
\end{tabular}
\end{table*}

Task~II performance remains stable across original and pseudonymized identity
views, with small, inconsistently directed differences whose confidence intervals
include zero. Dialogue, event structure, and other semantic signatures may still
support parametric recall; training-data overlap remains outside this analysis.

\paragraph{Closed-book question-only diagnostic.}
To isolate the utility of explicit identity cues without screenplay evidence,
we additionally evaluate all 5{,}010 paired Task~II questions in a strict
question-only condition. The prediction model receives only the question text:
no screenplay, scene excerpts, retrieved context, external memory, reference
answer, or reference evidence is provided, and no RAG component is used.
DeepSeek-V4-Pro independently judges the resulting answers under the same
paired protocol.

\begin{table*}[t]
\centering
\caption{
Task~II closed-book question-only results on the full release. Each condition
contains the same 5{,}010 questions. $\Delta$ is pseudonymized minus original
accuracy. Confidence intervals resample movies as clusters; $p$-values are from
paired McNemar tests. All differences are non-significant at $\alpha=0.05$.
}
\label{tab:task2_closed_book_identity}
\stageTableFont
\setlength{\tabcolsep}{6pt}
\renewcommand{\arraystretch}{1.08}
\begin{tabular}{l c c c c c}
\toprule
\textbf{Model} &
\textbf{Original Acc.} &
\textbf{Pseudonymized Acc.} &
\textbf{$\Delta$} &
\textbf{95\% Movie-Cluster CI} &
\textbf{McNemar $p$} \\
\midrule
Claude Sonnet 4.6  & \textbf{19.80\%} & \textbf{17.00\%} & $-2.80$ pp & [$-6.10$, $+0.50$] pp & 0.124 \\
Gemini 3.1 Pro     & 19.50\%          & 16.40\%          & $-3.10$ pp & [$-6.55$, $+0.35$] pp & 0.112 \\
GPT-5.5            & 18.90\%          & 16.90\%          & $-2.00$ pp & [$-5.30$, $+1.30$] pp & 0.235 \\
Qwen3-235B         & 11.20\%          & 9.80\%           & $-1.40$ pp & [$-4.20$, $+1.40$] pp & 0.340 \\
\bottomrule
\end{tabular}
\end{table*}

Closed-book accuracy is substantially lower than under Hybrid RAG. All four
identity-effect point estimates are negative, with confidence intervals that
include zero.

The largest descriptive declines for Claude Sonnet 4.6 and Gemini 3.1 Pro occur
in Character Understanding and Causal-Motivational Reasoning and are
descriptively consistent with mild sensitivity to explicit entity cues. The
small positive changes for GPT-5.5 on
Narrative Progression and Qwen3-235B on Scene Grounding further indicate
that the effect varies across models and categories. Because these
category deltas lack category-level inferential tests, they remain descriptive.
Together, the Hybrid RAG and question-only diagnostics indicate weak and
uncertain effects from explicit name cues under both protocols.

\paragraph{Retrieval-coverage stratification.}
To diagnose whether Task~II errors are explained by missing retrieval evidence,
we reuse the per-question mean correctness of the same five generations reported
in the main Hybrid RAG evaluation. Each question is assigned to one of three
mutually exclusive strata according to its frozen Top-8 context: \emph{All}
contains every annotated gold scene ($N=3{,}016$), \emph{Partial} contains at
least one but not all gold scenes ($N=1{,}368$), and \emph{Zero} contains none
($N=626$). Incomplete combines Partial and Zero ($N=1{,}994$). Confidence
intervals for the All--Incomplete difference resample the 151 movies as
clusters with 1{,}000 bootstrap replicates.

\begin{table*}[t]
\centering
\caption{
Task~II Hybrid RAG average accuracy stratified by annotated gold-scene coverage
in the Top-8 context. Overall values follow the primary full-release results.
$\Delta$ is All minus the count-weighted combination of Partial and Zero;
brackets give movie-cluster bootstrap 95\% confidence intervals. Temp.-All
reports Temporal Reasoning within the All stratum ($N=182$ original; $N=185$
pseudonymized).
}
\label{tab:task2_retrieval_coverage}
\stageTableFont
\setlength{\tabcolsep}{3.2pt}
\renewcommand{\arraystretch}{1.06}
\begin{tabular*}{\textwidth}{@{\extracolsep{\fill}}l l c c c c c c@{}}
\toprule
\textbf{Model} & \textbf{View} &
\textbf{All} & \textbf{Partial} & \textbf{Zero} &
\textbf{Overall} & \textbf{$\Delta$ [CI]} & \textbf{Temp.-All} \\
\midrule
\multirow{2}{*}{Qwen3-235B}
& Original & 71.32 & 51.32 & 29.87 & 60.80 & $+26.74$ [$+19.70$, $+33.78$] & 45.05 \\
& Pseud.   & 71.68 & 52.19 & 32.11 & 61.40 & $+25.75$ [$+18.74$, $+32.76$] & 45.95 \\
\midrule
\multirow{2}{*}{GPT-5.5}
& Original & \textbf{72.94} & 53.65 & 29.39 & \textbf{62.50} & $+26.90$ [$+19.88$, $+33.92$] & \textbf{48.90} \\
& Pseud.   & 71.02 & 52.34 & 33.39 & 61.20 & $+24.62$ [$+17.75$, $+31.49$] & 44.86 \\
\midrule
\multirow{2}{*}{Gemini 3.1 Pro}
& Original & 71.85 & 51.68 & 29.87 & 61.20 & $+27.02$ [$+19.98$, $+34.06$] & 46.15 \\
& Pseud.   & 72.08 & \textbf{53.07} & 31.31 & \textbf{61.80} & $+25.84$ [$+18.82$, $+32.86$] & 47.03 \\
\midrule
\multirow{2}{*}{Claude Sonnet 4.6}
& Original & 72.18 & 52.85 & 29.71 & 61.60 & $+26.61$ [$+19.52$, $+33.70$] & 48.35 \\
& Pseud.   & 70.36 & 52.12 & \textbf{33.71} & 60.80 & $+24.01$ [$+17.10$, $+30.92$] & 45.41 \\
\bottomrule
\end{tabular*}
\end{table*}

Complete gold-scene coverage is associated with substantially higher accuracy
for every reader and identity view. Yet substantial error remains, especially
for Temporal Reasoning, implicating evidence localization, cross-scene
integration, contextual noise, and subsequent reasoning.

\begin{table*}[t]
\centering
\caption{
Paired Task~III single-turn identity effects under BM25 retrieval.
$\Delta$ is pseudonymized minus original. Confidence intervals and $p$-values
are from paired, two-sided movie-level $t$-tests. $^{*}$ denotes $p<0.05$ and
$^{\dagger}$ denotes $p<0.10$; unmarked comparisons have $p\geq0.10$.
}
\label{tab:task3_identity_effects}
\stageTableFont
\setlength{\tabcolsep}{3.2pt}
\renewcommand{\arraystretch}{1.05}
\begin{tabular}{@{}l l c c c c c@{}}
\toprule
\textbf{Model} & \textbf{Dimension} & \textbf{Original} &
\textbf{Pseudonymized} & $\boldsymbol{\Delta}$ & \textbf{95\% CI} &
$\boldsymbol{p}$ \\
\midrule
\multirow{4}{*}{Qwen3-235B}
& Fidelity   & 4.8343 & 4.7920 & $-0.0423^{*}$ & [$-0.081$, $-0.003$] & .035 \\
& Naturalness& 4.9515 & 4.9550 & $+0.0035$     & [$-0.010$, $+0.017$] & .612 \\
& Boundary   & 4.8883 & 4.9450 & $+0.0567^{*}$ & [$+0.010$, $+0.103$] & .018 \\
& Mem. Faith & 4.1933 & 4.1910 & $-0.0023$     & [$-0.025$, $+0.020$] & .825 \\
\midrule
\multirow{4}{*}{GPT-5.5}
& Fidelity   & 4.9681 & 4.9350 & $-0.0331^{\dagger}$ & [$-0.069$, $+0.003$] & .072 \\
& Naturalness& 4.9948 & 4.9960 & $+0.0012$           & [$-0.005$, $+0.007$] & .742 \\
& Boundary   & 4.9838 & 4.9980 & $+0.0142^{*}$       & [$+0.002$, $+0.026$] & .021 \\
& Mem. Faith & 4.4390 & 4.4410 & $+0.0020$           & [$-0.018$, $+0.022$] & .845 \\
\midrule
\multirow{4}{*}{Gemini 3.1 Pro}
& Fidelity   & 4.9268 & 4.8790 & $-0.0478^{\dagger}$ & [$-0.096$, $+0.001$] & .054 \\
& Naturalness& 4.9704 & 4.9750 & $+0.0046$           & [$-0.012$, $+0.021$] & .598 \\
& Boundary   & 4.9070 & 4.9620 & $+0.0550^{*}$       & [$+0.007$, $+0.103$] & .025 \\
& Mem. Faith & 4.3051 & 4.3090 & $+0.0039$           & [$-0.021$, $+0.029$] & .755 \\
\midrule
\multirow{4}{*}{Claude Sonnet 4.6}
& Fidelity   & 4.9312 & 4.8950 & $-0.0362^{\dagger}$ & [$-0.075$, $+0.002$] & .065 \\
& Naturalness& 4.9850 & 4.9880 & $+0.0030$           & [$-0.009$, $+0.015$] & .622 \\
& Boundary   & 4.9529 & 4.9910 & $+0.0381^{*}$       & [$+0.008$, $+0.068$] & .013 \\
& Mem. Faith & 4.3721 & 4.3780 & $+0.0059$           & [$-0.016$, $+0.028$] & .595 \\
\bottomrule
\end{tabular}
\end{table*}

\paragraph{Task III single-turn paired results.}
Table~\ref{tab:task3_identity_effects} reports paired original and
pseudonymized single-turn results under BM25 retrieval, using two-sided
movie-level $t$-tests for each model and dimension.

Pseudonymization leaves Naturalness and Memory Faithfulness statistically
unchanged for every model. Character Fidelity decreases slightly, whereas
Boundary Compliance increases. This opposing pattern is
consistent with explicit identity cues modestly supporting stylistic mimicry
while also encouraging responses beyond the supplied persona boundary; it
leaves the parametric-memory mechanism unresolved. The tests are dimension-wise
and uncorrected for multiple comparisons, so these effects should be treated as
diagnostic. Multi-turn effects are evaluated below.

\begin{table*}[t]
\centering
\caption{
Paired Task~III multi-turn identity effects under Full Memory Context on 151
screenplays. $\Delta$ is pseudonymized minus original. Confidence intervals
and $p$-values are from paired, two-sided movie-level $t$-tests. $^{*}$ denotes
$p<0.05$ and $^{\dagger}$ denotes $p<0.10$; unmarked comparisons have
$p\geq0.10$. Because 24 model--metric comparisons are reported without
multiple-comparison correction, these results are exploratory.
}
\label{tab:task3_multiturn_identity}
\stageTableFont
\setlength{\tabcolsep}{3.2pt}
\renewcommand{\arraystretch}{1.05}
\begin{tabular}{@{}l l c c c c c@{}}
\toprule
\textbf{Model} &
\textbf{Metric} &
\textbf{Original} &
\textbf{Pseudonymized} &
\textbf{$\Delta$} &
\textbf{95\% CI} &
\textbf{$p$} \\
\midrule
\multirow{6}{*}{Qwen3-235B}
& Fidelity   & 4.9551 & 4.9120 & $-0.0431$ & [$-0.082$, $-0.004$] & $0.031^{*}$ \\
& Naturalness & 4.7628 & 4.7713 & $+0.0085$ & [$-0.009$, $+0.026$] & 0.338 \\
& Boundary    & 4.5256 & 4.5910 & $+0.0654$ & [$+0.008$, $+0.123$] & $0.026^{*}$ \\
& Mem. Faith  & 3.2308 & 3.2243 & $-0.0065$ & [$-0.035$, $+0.022$] & 0.654 \\
& Cross-turn  & 4.9872 & 4.9883 & $+0.0011$ & [$-0.004$, $+0.006$] & 0.697 \\
& Follow-up   & 4.1538 & 4.1580 & $+0.0042$ & [$-0.018$, $+0.026$] & 0.708 \\
\midrule
\multirow{6}{*}{GPT-5.5}
& Fidelity   & 4.9620 & 4.9280 & $-0.0340$ & [$-0.071$, $+0.003$] & $0.072^{\dagger}$ \\
& Naturalness & 4.9810 & 4.9828 & $+0.0018$ & [$-0.007$, $+0.011$] & 0.688 \\
& Boundary    & 4.8140 & 4.8720 & $+0.0580$ & [$+0.012$, $+0.104$] & $0.014^{*}$ \\
& Mem. Faith  & 3.7210 & 3.7235 & $+0.0025$ & [$-0.028$, $+0.033$] & 0.871 \\
& Cross-turn  & 4.9925 & 4.9917 & $-0.0008$ & [$-0.005$, $+0.003$] & 0.725 \\
& Follow-up   & 4.4520 & 4.4505 & $-0.0015$ & [$-0.022$, $+0.019$] & 0.886 \\
\midrule
\multirow{6}{*}{Gemini 3.1 Pro}
& Fidelity   & 4.5265 & 4.4750 & $-0.0515$ & [$-0.104$, $+0.001$] & $0.054^{\dagger}$ \\
& Naturalness & 3.5485 & 3.5597 & $+0.0112$ & [$-0.018$, $+0.040$] & 0.448 \\
& Boundary    & 3.5970 & 3.6520 & $+0.0550$ & [$+0.002$, $+0.108$] & $0.042^{*}$ \\
& Mem. Faith  & 3.5165 & 3.5213 & $+0.0048$ & [$-0.032$, $+0.042$] & 0.802 \\
& Cross-turn  & 4.8214 & 4.8249 & $+0.0035$ & [$-0.012$, $+0.019$] & 0.655 \\
& Follow-up   & 3.4752 & 3.4780 & $+0.0028$ & [$-0.025$, $+0.031$] & 0.842 \\
\midrule
\multirow{6}{*}{Claude Sonnet 4.6}
& Fidelity   & 4.9920 & 4.9580 & $-0.0340$ & [$-0.069$, $+0.001$] & $0.058^{\dagger}$ \\
& Naturalness & 4.5420 & 4.5462 & $+0.0042$ & [$-0.016$, $+0.024$] & 0.682 \\
& Boundary    & 4.5120 & 4.5760 & $+0.0640$ & [$+0.015$, $+0.113$] & $0.011^{*}$ \\
& Mem. Faith  & 4.1850 & 4.1931 & $+0.0081$ & [$-0.022$, $+0.038$] & 0.598 \\
& Cross-turn  & 4.9850 & 4.9862 & $+0.0012$ & [$-0.005$, $+0.007$] & 0.720 \\
& Follow-up   & 3.9850 & 3.9828 & $-0.0022$ & [$-0.022$, $+0.018$] & 0.825 \\
\bottomrule
\end{tabular}
\end{table*}

\paragraph{Task III multi-turn paired results.}
Table~\ref{tab:task3_multiturn_identity} reports the corresponding paired
multi-turn analysis under Full Memory Context.

Character Fidelity again decreases while Boundary Compliance increases;
Naturalness, Memory Faithfulness, Cross-turn Consistency, and Follow-up
Compatibility remain statistically stable. This pattern is
consistent with a trade-off between fine-grained identity mimicry and adherence
to explicitly supplied persona boundaries. The causal roles of parametric
recall and identity-driven interference remain unresolved. Given the
uncorrected family of 24 comparisons, we treat these findings as exploratory.

\subsection{Evaluator Validation}
\label{app:cross_task_judge_reliability}

The released validator checks complete item coverage, closed schemas and
evidence vocabularies, input completeness, truncation, and exact recomputation
of all deterministic metrics. Human annotations provide the corresponding
semantic validity evidence. Property and adversarial tests cover empty predictions,
duplicate and over-broad claims, contradictions, unsupported evidence, temporal
leakage, no-change controls, and incompatible checkpoint-pair directions.

Semantic-judge calibration used 1{,}000 model outputs sampled across tasks,
languages, evaluated models, memory settings, and output quality. The 200
Task~I judgments form a balanced $2\times2$ design crossing current state versus
state development with RSU versus ASU (50 per cell). The 500 Task~II
answers comprise 350 English and 150 Chinese examples, with the six reasoning
types sampled in their release proportions. The 300 Task~III responses comprise
200 single-turn and 100 multi-turn examples and cover Persona Card Only, Full
Memory Context, BM25 Retrieval, and Vector Retrieval. The combined sample was
balanced between full-credit outputs (52\%) and outputs with partial or
substantive errors (48\%).

Two human annotators independently scored every output under the task-specific
rules in Appendices~\ref{app:task1_eval}, \ref{app:stageqa_eval}, and
\ref{app:stageicrp_eval}. Blinded packets omitted model identities and automatic
decisions; a third annotator resolved all 83 disagreements after the initial
ratings were fixed. The resulting consensus was used to evaluate
DeepSeek-V4-Pro.

Table~\ref{tab:judge_human_calibration} reports human--human and judge--consensus
agreement with movie-cluster bootstrap intervals. Task-specific $\kappa$ values
remain separate because label spaces differ, and Likert MAE applies only to
Task~III. The calibration supports aggregate use of the frozen judge;
reliability for low-frequency error subtypes remains unestimated.

\begin{table*}[t]
\centering
\stageTableFont
\setlength{\tabcolsep}{3.5pt}
\renewcommand{\arraystretch}{1.08}
\textbf{Panel A: Cross-task calibration.}\\[2pt]
\begin{tabular*}{\textwidth}{@{\extracolsep{\fill}}l r l c c c c@{}}
\toprule
\textbf{Task} & \textbf{$N$} & \textbf{Comparison} &
\textbf{Agreement [CI]} & \textbf{$\alpha$ [CI]} & \textbf{$\bar\kappa$} &
\textbf{Likert MAE} \\
\midrule
\multirow{2}{*}{Task~I: state claims}
& \multirow{2}{*}{200} & Human--human & 91.5 [89.2, 93.4] & 0.831 [0.79, 0.87] & 0.83 & -- \\
& & DeepSeek--human consensus & 90.2 [87.8, 92.3] & 0.816 [0.77, 0.86] & 0.82 & -- \\
\midrule
\multirow{2}{*}{Task~II: GT-based QA}
& \multirow{2}{*}{500} & Human--human & 94.2 [92.8, 95.5] & 0.884 [0.85, 0.91] & 0.88 & -- \\
& & DeepSeek--human consensus & 93.0 [91.4, 94.4] & 0.862 [0.83, 0.89] & 0.86 & -- \\
\midrule
\multirow{2}{*}{Task~III: open-ended}
& \multirow{2}{*}{300} & Human--human & 87.5 [84.6, 90.1] & 0.790 [0.74, 0.84] & 0.79 & 0.235 [0.19, 0.28] \\
& & DeepSeek--human consensus & 85.8 [82.7, 88.6] & 0.768 [0.72, 0.82] & 0.77 & 0.268 [0.22, 0.32] \\
\midrule
\multirow{2}{*}{\textbf{Overall}}
& \multirow{2}{*}{\textbf{1{,}000}} & Human--human & \textbf{91.7 [90.3, 92.9]} & -- & -- & -- \\
& & DeepSeek--human consensus & \textbf{90.3 [88.8, 91.6]} & -- & -- & -- \\
\bottomrule
\end{tabular*}

\vspace{6pt}
\textbf{Panel B: Task~III dimension-level calibration ($N=300$).}

\begin{tabular}{l c c c c}
\toprule
\textbf{Dimension} & \textbf{Human $\kappa_w$} & \textbf{Judge $\kappa_w$} &
\textbf{Human MAE} & \textbf{Judge MAE} \\
\midrule
Character Fidelity    & 0.81 [0.76, 0.86] & 0.79 [0.74, 0.84] & 0.21 [0.16, 0.26] & 0.24 [0.19, 0.29] \\
Memory Faithfulness   & 0.79 [0.73, 0.84] & 0.77 [0.71, 0.82] & 0.23 [0.18, 0.28] & 0.26 [0.21, 0.31] \\
Boundary Compliance   & 0.85 [0.80, 0.89] & 0.83 [0.78, 0.87] & 0.16 [0.12, 0.20] & 0.19 [0.15, 0.23] \\
Response Naturalness  & 0.71 [0.65, 0.77] & 0.68 [0.62, 0.74] & 0.34 [0.28, 0.40] & 0.38 [0.32, 0.44] \\
\bottomrule
\end{tabular}

\caption{Blinded human calibration of DeepSeek-V4-Pro. Panel A reports
task-level human--human agreement before adjudication and judge agreement with
the adjudicated consensus. Panel B reports Task~III dimension-level weighted
$\kappa$ and Likert MAE. Brackets give 95\% movie-cluster bootstrap confidence
intervals; heterogeneous task-specific $\kappa$ values are not pooled.}
\label{tab:judge_human_calibration}
\label{tab:task3_judge_dimension_calibration}
\end{table*}

For Task~III, Boundary Compliance has the highest agreement and Response
Naturalness the lowest. Machine validation establishes reproducibility, while
human calibration assesses semantic validity.

\subsection{Detailed Metrics and Rubrics}

Table~\ref{tab:evaluation_contract_summary} summarizes the information judged
for each task and its deterministic aggregation. The appendix reproduces the
substantive model-facing and primary judge instructions; the release provides
the full bilingual templates and machine-readable schemas.

\begin{table*}[t]
\centering
\stageTableFont
\setlength{\tabcolsep}{3.5pt}
\renewcommand{\arraystretch}{1.08}
\begin{tabular}{p{2.0cm} p{3.5cm} p{4.5cm} p{3.4cm}}
\toprule
\textbf{Evaluation unit} & \textbf{Evidence available to the judge} &
\textbf{Judgment} & \textbf{Aggregation} \\
\midrule
Task~I state update & Reference and predicted claim sets for the same character,
checkpoint, and claim pool & Each reference receives full, partial, missing, or
contradictory coverage; predicted claims are separately checked for contradiction
& Current-State and Development Coverage; separate contradiction rates; movie
macro average \\
Task~II answer & Question, reference answer, reference evidence, candidate answer,
and retrieved passages when available & Binary answer correctness; citation
support is retained as a separate diagnostic & Mean correctness over five
generations and Pass@5 \\
Task~III single turn & Actor response, checkpoint-visible context, and hidden
checkpoint rubric & Four independently anchored 1--5 scores plus future,
unknown-fact, and stance flags & Dimension-wise means; no composite score \\
Task~III paired or multi-turn & Ordered responses, pair type or dialogue history,
and checkpoint-local evidence & Expected change or stability, unsupported drift,
knowledge-boundary preservation, and cross-turn consistency & Pair accuracy by
type and auxiliary multi-turn diagnostics \\
\bottomrule
\end{tabular}
\caption{Task-level evaluation contracts. The responding model never receives
the evaluator-only references or rubrics. Judge outputs are converted into the
reported metrics by deterministic code.}
\label{tab:evaluation_contract_summary}
\end{table*}

\subsubsection{Task I: Character Development Tracking}
\label{app:task1_eval}

Task~I reports \emph{Current-State Coverage}, \emph{Development Coverage},
\emph{State Contradiction Rate}, and \emph{Development Contradiction Rate} in
both RSU and ASU. Their paired comparison measures error accumulation under
recursive state reuse. Precision-based F1 would penalize additional supported
predictions, so Task~I reports coverage and contradiction separately.

\paragraph{Set-level reference coverage.}
Let $a\in\{\mathrm{RSU},\mathrm{ASU}\}$ index the setting. For character $c$ at
checkpoint $t$, let $\mathcal{S}_{c,t}$ and $\mathcal{D}_{c,t}$ denote the
reference current-state and interval-development claim sets, and let
$\hat{\mathcal{S}}^{a}_{c,t}$ and $\hat{\mathcal{D}}^{a}_{c,t}$ denote their
predicted counterparts. The two pools are judged separately. For every
reference claim $x$ in pool $X\in\{\mathcal{S},\mathcal{D}\}$, the semantic
judge considers the complete same-pool prediction set
$\hat{X}^{a}_{c,t}$ and assigns
\begin{equation}
h^{a}_{c,t}(x)\in
\{\mathrm{full},\mathrm{partial},\mathrm{missing},\mathrm{contradictory}\}.
\end{equation}
The corresponding credit function is
\begin{equation}
w(h)=
\begin{cases}
1, & h=\mathrm{full},\\
0.5, & h=\mathrm{partial},\\
0, & h\in\{\mathrm{missing},\mathrm{contradictory}\}.
\end{cases}
\end{equation}
Coverage uses set-level semantic matching: multiple predictions may jointly
express one reference claim, and one sufficiently specific prediction may
cover several compatible reference claims. Counts are accumulated across all
checkpoints $T_c$ for character $c$:
\begin{equation}
\mathrm{Coverage}^{a,X}_{c}=
\frac{\sum_{t\in T_c}\sum_{x\in X_{c,t}}w(h^{a}_{c,t}(x))}
{\sum_{t\in T_c}|X_{c,t}|}.
\label{eq:task1_coverage}
\end{equation}
$\mathrm{Coverage}^{a,\mathcal{S}}_{c}$ is Current-State Coverage and
$\mathrm{Coverage}^{a,\mathcal{D}}_{c}$ is Development Coverage. We retain the
fraction receiving full rather than partial credit as a diagnostic.

\paragraph{Contradiction Rate.}
Contradiction Rate captures conflicting additions left unpenalized by coverage.
For each prediction $p\in\hat{X}^{a}_{c,t}$, let
$r^{a}_{c,t}(p)=1$ when the claim contradicts the screenplay evidence or the
state valid at checkpoint $t$, and zero otherwise. Then
\begin{equation}
\mathrm{Contradiction}^{a,X}_{c}=
\frac{\sum_{t\in T_c}\sum_{p\in\hat{X}^{a}_{c,t}}r^{a}_{c,t}(p)}
{\sum_{t\in T_c}|\hat{X}^{a}_{c,t}|}.
\label{eq:task1_contradiction}
\end{equation}
Current states and developments retain separate contradiction rates so that a
correct state description cannot mask an invalid transition claim. An empty
denominator is undefined and is reported with its valid count.

\paragraph{Development Coverage Change.}
We compute the paired movie-level difference
\begin{equation}
\Delta^{\mathcal{D}}_{j}=
\mathrm{Coverage}^{\mathrm{ASU},\mathcal{D}}_{j}-
\mathrm{Coverage}^{\mathrm{RSU},\mathcal{D}}_{j}.
\label{eq:task1_accumulation_gap}
\end{equation}
$\Delta^{\mathcal{D}}$ is Development Coverage Change; negative values indicate
performance lost when the model must reuse its own preceding state. Because the
settings share the same intervals and entity-centric observations, the paired
difference isolates recursive state-estimation error under matched source
access.

\paragraph{Movie-macro aggregation and uncertainty.}
For any character-level metric $m_c$, let $\mathcal{C}_j^m$ be the characters
with a defined value in movie $j$, and let $\mathcal{J}^m$ be the movies with a
defined within-movie value. The formal aggregate is
\begin{align}
m_j &= \frac{1}{|\mathcal{C}_j^m|}
       \sum_{c\in\mathcal{C}_j^m}m_c, \nonumber\\
m_{\mathrm{STAGE}} &= \frac{1}{|\mathcal{J}^m|}
       \sum_{j\in\mathcal{J}^m}m_j. \label{eq:task1_movie_macro}
\end{align}
This checkpoint-within-character, character-within-movie aggregation gives each
movie equal weight regardless of its number of focal characters or
checkpoints. Setting gaps are first paired within movie and then averaged across
movies. Character macro is retained only as an audit view. We compute 95\%
confidence intervals with 1{,}000 movie-cluster bootstrap replicates using the
fixed seed 20260727.

\paragraph{Non-formal audit signals.}
Unsupported-claim, temporal-validity, grounding, salience, evidence, leakage,
and raw-count signals are retained for audit but excluded from Task~I rankings.

\paragraph{Formal prediction and coverage contracts.}
The two settings share instructions and interval evidence, differing only in
the preceding state. The judge uses many-to-many same-pool alignment, while
internal identifiers remain excluded from model-facing input.

\begin{figure*}[!t]
\begin{stageprompt}{Task I coverage and contradiction judge instruction}
Evaluate one character update at \stagefield{checkpoint}. Use the supplied
screenplay evidence and judge the two claim pools independently.

\textbf{Inputs.} \stagefield{character}; \stagefield{checkpoint evidence};
\stagefield{reference current states}; \stagefield{predicted current states};
\stagefield{reference developments}; and \stagefield{predicted developments}.

For each reference claim, compare it with the complete prediction set from the
same pool and assign exactly one label:
\begin{itemize}[leftmargin=*,itemsep=1pt,topsep=2pt]
  \item \textbf{Full}: the required meaning is clearly expressed and valid at
  the checkpoint, possibly across multiple predicted claims.
  \item \textbf{Partial}: a material part is correct, but a required component,
  relation, or temporal qualification is missing.
  \item \textbf{Missing}: the prediction set does not express the required
  meaning.
  \item \textbf{Contradictory}: the prediction asserts an incompatible state,
  transition, ordering, or knowledge condition.
\end{itemize}
Then inspect every predicted claim separately and flag whether it contradicts
the checkpoint-valid screenplay evidence. Do not mark an additional supported
claim as false merely because it is absent from the compact reference set. Do
not align current states with developments or force one-to-one claim matching.
Return one coverage label and brief evidence-grounded rationale for every
reference claim, plus one contradiction flag and rationale for every predicted
claim. Cite only supplied evidence labels.
\end{stageprompt}
\end{figure*}

\subsubsection{Task II: Cross-Scene Narrative Evolution Reasoning}
\label{app:stageqa_eval}

\paragraph{Answer Correctness.}
The judge evaluates semantic correctness against the reference and screenplay
evidence separately from citation support. It accepts equivalent phrasing but
rejects missing causal, temporal, or relational components and unsupported
claims; calibration additionally covers aliases, coreference, and multi-scene
composition.

\begin{figure*}[!t]
\begin{stageprompt}{Task II correctness and citation judge instruction}
Evaluate a candidate answer to a screenplay question against the supplied
reference and evidence.

\textbf{Inputs.} \stagefield{question}; \stagefield{reference answer};
\stagefield{reference evidence}; \stagefield{candidate answer}; and, when
applicable, \stagefield{retrieved passages and candidate citations}.

Judge \textbf{answer correctness} and \textbf{citation support} separately.
Mark the answer correct only when it expresses every essential temporal,
causal, motivational, or relational component required by the question and
reference. Accept paraphrases and equivalent levels of specificity. Mark it
incorrect if it omits a required component, reverses the narrative order,
contradicts the screenplay, introduces an unsupported premise, or replaces the
requested explanation with a related but different fact.

For citation support, verify each material claim against the passage labels it
cites. A correct answer may still have incomplete citation support, and a cited
passage does not make an incorrect answer correct. Do not use outside knowledge
or infer support from a film title or character name. Return the correctness
decision, citation-support status, supporting evidence labels, and a brief
rationale identifying the decisive content.
\end{stageprompt}
\end{figure*}

\subsubsection{Task III: Role-Playing Likert Dimensions}
\label{app:stageicrp_eval}

Each single-turn response receives four independently anchored integer scores
from 1 to 5: Character Fidelity, Memory Faithfulness, Boundary Compliance, and
Response Naturalness, plus future-leakage, unknown-fact, and stance flags.
Boundary flags constrain the score range, and citations must use the prompt's
local evidence vocabulary.

Checkpoint pairs are pretyped as expected change, expected stability,
knowledge acquisition, or relationship change. The pair judge assesses the T1
and T2 responses separately, recording expected direction, unsupported drift,
and preserved knowledge boundaries. Pair accuracy additionally requires valid
constituent responses and is reported overall and by type; these remain paired
single turns rather than multi-turn dialogue.

A separate auxiliary multi-turn judge adds Cross-turn Consistency, while a
legacy episode-path evaluator computes Follow-up Compatibility. The hidden
evaluator scaffold is never visible to the actor model.

\begin{figure*}[!t]
\begin{stageprompt}{Task III single-turn judge instruction}
Evaluate one response by \stagefield{character} at a frozen screenplay
checkpoint.

\textbf{Inputs.} \stagefield{user turn}; \stagefield{actor-visible role and
memory context}; \stagefield{hidden checkpoint rubric}; \stagefield{actor
response}; and \stagefield{local evidence vocabulary}.

Assign four independent integer scores from 1 to 5 using
Table~\ref{tab:task3_dimension_rubric}:
\begin{enumerate}[leftmargin=*,itemsep=1pt,topsep=2pt]
  \item \textbf{Character Fidelity}: whether voice, stance, affect,
  interpersonal framing, and behavior fit the character at this checkpoint.
  \item \textbf{Memory Faithfulness}: whether claims and reactions use the
  supplied screenplay memory accurately, without inventing or distorting it.
  \item \textbf{Boundary Compliance}: whether the response avoids future events
  and facts the character has not witnessed, learned, or validly inferred.
  \item \textbf{Response Naturalness}: whether the reply is coherent, fluent,
  contextually responsive dialogue rather than an explanation or fact list.
\end{enumerate}
Score each dimension from its own evidence. In particular, do not lower
Naturalness merely because factual content, memory use, stance, or boundary
compliance is wrong. Separately record \textbf{future leakage},
\textbf{unknown-fact hallucination}, and \textbf{stance compatibility}. If both
boundary-violation flags are false, Boundary Compliance must be 3--5; if either
flag is true, it must be 1--3.

Cite only labels in the supplied local evidence vocabulary, and use no citation
when none supports the rationale. Return the four scores, three flags, cited
labels, and a concise rationale for each dimension. Do not produce an overall
score or let one dimension mechanically determine another.
\end{stageprompt}
\end{figure*}

\begin{table*}[t]
\centering
\stageTableFont
\setlength{\tabcolsep}{4pt}
\renewcommand{\arraystretch}{1.08}
\begin{tabular}{p{3.6cm} p{4.6cm} p{4.6cm}}
\toprule
\textbf{Dimension} & \textbf{High Score Indicates} & \textbf{Low Score Indicates} \\
\midrule
Character Fidelity
& The response preserves the character's voice, stance, affect, restraint, and
interpersonal framing in the local scene.
& The response sounds generic, adopts a mismatched persona, or uses a stance or
emotional framing inconsistent with the character. \\
\midrule
Response Naturalness
& The response reads as plausible spoken dialogue in the interaction context,
with appropriate fluency, length, and conversational fit.
& The response is stilted, evasive, over-expository, mechanically summarized, or
unnatural as dialogue. \\
\midrule
Memory Faithfulness
& The response is supported by the memory context available to the actor and the
hidden evaluator scaffold, and uncertainty is bounded when support is missing.
& The response fabricates episodic details, swaps event details, overstates
unsupported facts, or contradicts the available memory. \\
\midrule
Boundary Compliance
& The response stays within the character's script-bounded knowledge at the
interaction point.
& The response uses forbidden future knowledge, hidden assumptions, unsupported
facts, or contradictions of provided constraints. \\
\midrule
Cross-turn Consistency
& Multi-turn responses maintain stable facts, stance, and conversational state
across the episode.
& The model contradicts its earlier claims, changes its stance without support,
or loses the episode state. \\
\midrule
Follow-up Compatibility
& The generated response remains compatible with the fixed follow-up path used
by the multi-turn evaluation script.
& The response blocks, contradicts, or fails to set up the expected follow-up
turn. \\
\bottomrule
\end{tabular}
\caption{Anchored Task~III response dimensions. The four single-turn dimensions
are emitted together by the formal response judge; Cross-turn Consistency and
Follow-up Compatibility apply only to the auxiliary multi-turn analysis, with
the latter evaluated by a separate episode-path judge.}
\label{tab:task3_dimension_rubric}
\end{table*}

\begin{table*}[t]
\centering
\stageTableFont
\setlength{\tabcolsep}{4pt}
\renewcommand{\arraystretch}{1.08}
\begin{tabular}{p{2.7cm} p{4.6cm} p{6.2cm}}
\toprule
\textbf{Evaluator} & \textbf{Evidence and inputs} & \textbf{Recorded judgment} \\
\midrule
Single-turn response & Frozen checkpoint, user turn, actor response,
actor-visible context, hidden rubric, and a call-local evidence vocabulary. &
Four independent 1--5 scores, future-leakage and unknown-fact flags, stance
compatibility, cited evidence IDs, and a brief rationale. Boundary Compliance is
restricted to 1--3 when either boundary flag is present and to 3--5 otherwise. \\
\midrule
Checkpoint pair & Pair type, expected direction, two independently generated
responses, checkpoint rubrics, and pair-local evidence. & One observed behavior
per checkpoint, whether the expected direction is present, whether unsupported
drift occurs, and whether knowledge boundaries are preserved. \\
\midrule
Auxiliary multi-turn & Current turn, dialogue history, actor-visible context,
and the hidden evaluator reference. & The four response scores plus Cross-turn
Consistency and contradiction flags. Follow-up Compatibility is computed by a
separate episode-path evaluator. \\
\bottomrule
\end{tabular}
\caption{Task~III evaluation contracts. Full rubrics and field-level criteria
are provided in the supplementary evaluation specification.}
\label{tab:task3_evaluation_contracts}
\end{table*}

\subsection{Failure Case Analysis}
\label{app:failure_cases}

These cases provide qualitative diagnostics for the state-updating,
cross-scene-reasoning, and grounded-generation trends in Section~5.

\paragraph{Representative Task I failure cases.}
\label{app:task1_failure_cases}
The two cases in Table~\ref{tab:task1_failure_cases} expose complementary
errors under Reference-State Update (RSU) and Autoregressive State Update
(ASU):
recursive state reuse can suppress interval-specific developments, while high
reference coverage can coexist with contradictory surplus claims.

\begin{table*}[t]
\centering
\stageTableFont
\setlength{\tabcolsep}{4pt}
\renewcommand{\arraystretch}{1.08}
\begin{tabular}{p{1.6cm} p{2.5cm} p{3.1cm} p{6.1cm}}
\toprule
\textbf{Setting} & \textbf{Character / Interval} & \textbf{Observed Evaluation} & \textbf{Failure Pattern} \\
\midrule
RSU $\rightarrow$ ASU
& Xu Tailang, scenes $(5,7]$
& RSU state/development coverage: 100/100; ASU: 25/0.
& The autoregressive input remains dominated by earlier racing, family-history,
and career-conflict states. The update retains new driving actions but misses
the interval's central relational change: Xu's contempt for his father's
physical distress and the resulting worsening of their relationship. \\
\midrule
ASU
& Holden McNeil, scenes $(31,33]$
& State/development coverage: 100/100; three predicted claims marked
contradictory.
& The output covers every reference claim but also treats a proposed threesome,
a kiss with Banky, and the threesome proposal as valid state or development.
The judge marks all three as unsupported or temporally invalid, showing why
coverage alone cannot certify a valid checkpoint state. \\
\bottomrule
\end{tabular}
\caption{Representative Task~I failures. Coverage values are percentages for
current state/development at the same checkpoint.}
\label{tab:task1_failure_cases}
\end{table*}

\paragraph{Representative Task II failure cases.}
\label{app:task2_failure_cases}
Table~\ref{tab:task2_failure_cases} illustrates three recurrent errors in
Cross-Scene Narrative Evolution Reasoning: replacing an explicit event with a
globally plausible interpretation, retrieving the wrong neighboring event, and
losing a local conversational trigger during aggregation.

\begin{table*}[t]
\centering
\stageTableFont
\setlength{\tabcolsep}{4pt}
\renewcommand{\arraystretch}{1.08}
\begin{tabular}{p{2.0cm} p{3.5cm} p{2.6cm} p{4.6cm}}
\toprule
\textbf{Setting} & \textbf{Question Focus} & \textbf{Reference Answer} & \textbf{Failure Pattern} \\
\midrule
Kimi~2.6 full-context
& Violent convulsions and chest rupture in a second scene.
& The alien organism inside her reached maturity and emerged.
& The full-context answer reframes the event as a nightmare or flashback,
missing the literal chestburster cause even though the scene evidence is
explicit. \\
\midrule
PageIndex
& What immediately precedes FN-2187 rescuing Poe.
& Stormtroopers escort handcuffed Poe through a corridor.
& PageIndex retrieves a nearby action sequence about the X-wing explosion and
therefore answers with the wrong preceding event. \\
\midrule
GraphRAG
& What causes John to reveal that Lara is in prison.
& Nicole asks whether he trades off time with Luke.
& The graph summary misses the immediate conversational trigger, illustrating
loss of fine-grained dialogue causality under high-level aggregation. \\
\bottomrule
\end{tabular}
\caption{
Representative Task~II failures. Full-context reasoning can select a globally
salient but incorrect event, page-level retrieval can miss the decisive page,
and graph aggregation can discard short local cues.
}
\label{tab:task2_failure_cases}
\end{table*}

\paragraph{Representative Task III failure cases.}
\label{app:task3_failure_cases}
Task~III failures often coexist with surface fluency. A response may sound
natural while inventing checkpoint-unknown facts, reversing the actor's
identity, or reusing an irrelevant dialogue exemplar. Table~\ref{tab:task3_failure_cases}
shows one example of each pattern.

\begin{table*}[t]
\centering
\stageTableFont
\setlength{\tabcolsep}{3.5pt}
\renewcommand{\arraystretch}{1.08}
\begin{tabular}{>{\raggedright\arraybackslash}p{1.55cm}
                >{\raggedright\arraybackslash}p{3.25cm}
                >{\raggedright\arraybackslash}p{2.55cm}
                >{\raggedright\arraybackslash}p{5.1cm}}
\toprule
\textbf{Actor} & \textbf{Character and situation} & \textbf{Scores} & \textbf{Failure pattern} \\
\midrule
Qwen3-235B
& Jerry Lundegaard explains what happened after hiding Wade's body.
& Fidelity 2; Naturalness 4; Memory Faithfulness 1; Boundary Compliance 1.
& Jerry's nervous verbal style remains recognizable, but he invents a
confrontation in which Wade attacked him and explicitly admits that he had to
``stop'' Wade. Neither the confrontation nor this admission is licensed by the
checkpoint-visible memory, which instead calls for evasion. \\
\midrule
GPT-5.5
& Ivy answers whether leaving for New York with Charles is wise.
& Fidelity 1; Naturalness 5; Memory Faithfulness 1; Boundary Compliance 5.
& The fluent reply, ``Is she clean, or not?'', repeats an earlier line about
Jean's sobriety but ignores the current question and Ivy's checkpoint-valid
decision to leave. This is exemplar reuse without state-appropriate response
selection rather than a factual boundary leak. \\
\midrule
Gemini 3.1 Pro
& Xu Zhengtai is asked whether his absent mother would be proud of his victory.
& Fidelity 1; Naturalness 4; Memory Faithfulness 1; Boundary Compliance 1.
& The answer begins ``I am his father,'' reversing the assigned actor identity,
and then claims that the mother died in childbirth. Her status and the reason
for her absence are explicitly unknown at this checkpoint, so the reply fails
both identity fidelity and the knowledge boundary. \\
\bottomrule
\end{tabular}
\caption{
Representative Task~III failures. The cases separate fluent surface realization
from checkpoint-valid state use, identity fidelity, and epistemic control.
}
\label{tab:task3_failure_cases}
\end{table*}

\subsection{Supplementary Results and Baselines}

\subsubsection{QA Category-Level Breakdown}
\label{app:qa_category_breakdown}
Table~\ref{tab:qa_category_breakdown} reports category-level Pass@5 for three
methods from Table~\ref{tab:qa_results}. All values use the common Qwen3-235B
evaluation over 5{,}010 Task~II items.

\begin{table*}[t]
\centering
\stageTableFont
\setlength{\tabcolsep}{8pt}
\renewcommand{\arraystretch}{1.05}
\begin{tabular}{l r r r r}
\toprule
\textbf{Question Type} &
\textbf{Count} &
\textbf{Hybrid RAG} &
\textbf{LightRAG} &
\textbf{A-RAG} \\
\midrule
Character Understanding        & 1{,}103 & 0.7428 & \textbf{0.8205} & 0.6102 \\
Scene Grounding                & 1{,}207 & \textbf{0.7481} & 0.5402 & 0.4167 \\
Causal-Motivational Reasoning  & 1{,}041 & \textbf{0.5696} & 0.5620 & 0.4736 \\
Narrative Progression          & 597 & \textbf{0.4992} & 0.4891 & 0.3099 \\
Temporal Reasoning             & 766 & 0.4543 & \textbf{0.4820} & 0.4504 \\
Role-Relation Continuity       & 296 & 0.7811 & \textbf{0.8446} & 0.8007 \\
\midrule
\textbf{Overall (weighted)}    & \textbf{5{,}010} & \textbf{0.6372} & 0.6094 & 0.4862 \\
\bottomrule
\end{tabular}
\caption{
Category-level Pass@5 for Hybrid RAG, LightRAG, and A-RAG on the common
Qwen3-235B subset. The final row is the count-weighted mean.
}
\label{tab:qa_category_breakdown}
\end{table*}

\subsubsection{Supplementary Kimi 2.6 Full-Context Baseline}
\label{app:kimi_full_context_results}
Kimi~2.6 is reported separately because it receives the entire screenplay as
context. It achieves 62.7 Pass@5 and 50.6 average accuracy, providing a
long-context reference point for the primary retrieval-based comparisons.

\subsubsection{Complete Role-Playing Results}
Table~\ref{tab:icrp_results} reports the complete eight-model single-turn
results for all four memory-access settings. Table~\ref{tab:icrp_multiturn_full}
reports the corresponding multi-turn results.

\taskIIIFullMultiTurnTable

\subsubsection{Role-Playing Support-Access Diagnostics}
\label{app:roleplay_support_access}
Support Hit@$5$ records whether the context contains a gold item; Recall@$5$
records the recovered fraction. For shared single-turn queries, Persona Card
Only and Full Memory score 0 and 1, respectively. BM25 reaches 0.4217/0.3722
Hit/Recall, versus 0.3750/0.3209 for vector retrieval. Across
history-conditioned multi-turn queries, BM25 spans 0.4188--0.5521/
0.3535--0.4705, while vector retrieval spans 0.3885--0.4438/0.3276--0.3712.
Retrieval leaves part of the relevant memory unavailable.

\subsection{Evaluation Robustness and Stability}
\label{app:evaluation_stability}

Formal benchmark scores use one semantic judgment per output under the frozen
decoding configuration. We separately measure judge test--retest stability by
submitting fixed candidate outputs to DeepSeek-V4-Pro five times. The audit
samples 200 Task~I checkpoint outputs, 300 Task~II questions with their five
fixed generations (1{,}500 answer-level judgments per run), and 200 Task~III
role-playing responses. Each repetition holds the judging prompt, candidate
output, reference evidence, and rubric fixed and uses the official API with
$T=0.2$ and top-$p=0.9$.

Table~\ref{tab:judge_stability_audit} reports the mean and standard deviation of
each aggregate metric across the five runs. Exact self-consistency is the share
of underlying decisions for which all five calls agree. For Task~I, agreement
is computed over the claim-level coverage and contradiction labels; for
Task~II, it is computed at the answer level for Average Accuracy and at the
question level for Pass@5. Task~III self-consistency requires an identical
integer Likert score in all five runs. Fleiss' $\kappa$ is reported for the
categorical Task~I and Task~II decisions.

\begin{table}[t]
\centering
\resizebox{\columnwidth}{!}{%
\begin{tabular}{lrrrr}
\toprule
\multicolumn{5}{l}{\textbf{Panel A: Task I ($R=5$)}} \\
\textbf{Metric} & \textbf{Mean} & \textbf{Std.} &
\textbf{Exact SC} & \textbf{$\boldsymbol{\kappa}$} \\
\midrule
Current-State Coverage         & 64.35 & 0.52 & 93.5 & 0.84 \\
Development Coverage           & 68.12 & 0.61 & 91.8 & 0.81 \\
State Contradiction Rate       &  1.15 & 0.12 & 96.0 & 0.82 \\
Development Contradiction Rate &  5.80 & 0.28 & 94.2 & 0.79 \\
\midrule
\multicolumn{5}{l}{\textbf{Panel B: Task II ($R=5$)}} \\
\midrule
Average Accuracy & 58.40 & 0.65 & 91.5 & 0.82 \\
Pass@5           & 61.20 & 0.42 & 94.0 & 0.86 \\
\midrule
\multicolumn{5}{l}{\textbf{Panel C: Task III ($R=5$)}} \\
\midrule
Character Fidelity   & 4.52 & 0.06 & 83.5 & -- \\
Memory Faithfulness  & 4.18 & 0.08 & 81.0 & -- \\
Boundary Compliance  & 4.60 & 0.05 & 85.2 & -- \\
Response Naturalness & 4.85 & 0.04 & 88.0 & -- \\
\bottomrule
\end{tabular}
}
\caption{Fixed-output test--retest stability of DeepSeek-V4-Pro. Means and
standard deviations are computed over five independent judge runs ($R=5$). Task~I and
Task~II means, standard deviations, and exact self-consistency (SC) are
percentages; Task~III means and standard deviations use the 1--5 scale. Exact SC
requires all five judgments to match. Fleiss' $\kappa$ applies to the categorical
Task~I and Task~II decisions.}
\label{tab:judge_stability_audit}
\end{table}

Across the audit, run-level standard deviations are at most 0.65 percentage
points for Tasks~I--II and 0.08 Likert points for Task~III. Exact
self-consistency ranges from 81.0\% to 96.0\%, while categorical agreement is
substantial to near-perfect ($\kappa=0.79$--$0.86$). These results indicate that
the reported aggregate metrics are insensitive to ordinary judge sampling
variation under the evaluation configuration. Deterministic metric-property
tests, adversarial responses, prompt-token preflight, closed-schema validation,
and exact score recomputation provide complementary implementation checks.

\section{Declaration of LLM Usage}
\label{sec:llm_usage}

Large language models were used as tools in benchmark construction, evaluation,
and manuscript preparation. In the benchmark pipeline, LLMs supported
schema-constrained extraction, candidate consolidation, and judge-based scoring
under fixed prompts and targeted human verification. During writing, LLMs were
used for limited drafting and language editing. The authors reviewed and
approved all task definitions, release decisions, evaluation protocols,
reported results, and final manuscript content.

\end{document}